\documentclass[preprint,a4paper]{elsarticle}
\usepackage[margin=1.25in]{geometry}
\usepackage[english]{babel}
\usepackage{amsmath}
\usepackage{amsfonts}
\usepackage{graphicx}
\usepackage{subfig}
\usepackage{bm}
\usepackage{amssymb}
\usepackage{mathtools}
\usepackage{multirow}
\usepackage{subfig}
\biboptions{sort&compress}
\DeclareMathOperator{\diag}{diag}
\DeclareMathOperator*{\argminA}{arg\,min}

\title{%\vspace{-1.5cm}            % Another way to do
Constraining Gaussian processes for physics-informed acoustic emission mapping}

\author{M.R. Jones, T.J. Rogers, E.J. Cross}

\begin{document}

\maketitle
\makeatother

% \section*{To Do}

% \begin{enumerate}
%     \item Overall, need to think about work more strongly from the perspective of AE. That needs to be the main motivation, as opposed to here's a cool method, let's apply it to this. NEW FORMAT: I would talk about the importance of AE localisation, it's challenges and then say this can be met by these approaches and we are the first people to do it!
%     \item Motivate points in results and discussion at start - why they are interesting and useful problems for us to be solving, then here's this great method. 
%     \item Include other ways to incorporating boundaries earlier on (i.e. artificial measurements, and why these are not necessarily what we want all the time - particularly in the case of derivatives)
% \end{enumerate}

% and how by introcuing, it is possible to construct a model that can both learn from observed data whilst being physically meaningful

\section*{Abstract}

The automated localisation of damage in structures is a challenging but critical ingredient in the path towards predictive or condition-based maintenance of high value structures. The use of acoustic emission time of arrival mapping is a promising approach to this challenge, but is severely hindered by the need to collect a dense set of artificial acoustic emission measurements across the structure, resulting in a lengthy and often impractical data acquisition process. In this paper, we consider the use of physics-informed Gaussian processes for learning these maps to alleviate this problem. In the approach, the Gaussian process is constrained to the physical domain such that information relating to the geometry and boundary conditions of the structure are embedded directly into the learning process, returning a model that guarantees that any predictions made satisfy physically-consistent behaviour at the boundary. A number of scenarios that arise when training measurement acquisition is limited, including where training data are sparse, and also of limited coverage over the structure of interest. Using a complex plate-like structure as an experimental case study, we show that our approach significantly reduces the burden of data collection, where it is seen that incorporation of boundary condition knowledge significantly improves predictive accuracy as training observations are reduced, particularly when training measurements are not available across all parts of the structure. 

\bigskip

\noindent \textit{Keywords}: Physics-informed learning; Damage localisation; Acoustic emission; Constrained Gaussian processes

\section{Introduction}

%\begin{enumerate}
   % \item SHM intro and data-driven SHM, leading into grey-box
   % \item brief consideration of different approaches to grey-box
   % \item constrained machine learners as a means for physics informed ML 
   % \item constrained GPs
%\end{enumerate}

 %For example, data streams obtained from engineering infrastructure are often unlabelled, limiting the identification of damage to a problem of novelty detection. From this, one can only infer as to whether some change has occurred in the monitored features. Whilst this restriction is sufficient for the lower levels of the health monitoring framework such as damage detection, and in some restricted cases, damage localisation, moving to the upper parts of the hierarchy requires labelled data from multiple health states. Furthermore, purely data-driven tools should be viewed as no more than interpolators. In practice, this means that the data used to train a learning algorithm must cover the full range of operational and environmental conditions that the structure will be subject to. 

In the field of structural health monitoring (SHM), the objective is to implement monitoring strategies that seek to detect and assess damage that is present in engineering infrastructure \cite{farrar2012structural}. One particular branch of SHM techniques are those that consider a \textit{data-driven} perspective, treating the damage identification paradigm as a problem of pattern recognition. Under such an approach, the task is to first collect data from a structure of interest by means of sensing hardware. Features sensitive to damage are then extracted and used to learn a statistical model that can inform damage identification. 

Within the last decade or so, approaches that adopt such a viewpoint have become increasingly popular \cite{rogers2019bayesian,bull2018active,avendano2017gaussian,cross2012}. One area of SHM that has benefited from the use of data-driven approaches is damage localisation via acoustic emission (AE) \cite{tobias1976acoustic,hensman2010locating,ciampa2012impact,niri2014nonlinear,kundu2014acoustic}, enabling progression in scenarios that previously presented significant difficulty, such as for complex geometries or inhomogeneous material compositions. Whilst there are many physical mechanisms that can produce acoustic emission waves, AE activity can often be attributed to the formation and progression of damage. For example, as a crack begins to form and grow, the associated redistribution of internal stresses will cause a small amount of elastic energy to be released, manifesting as a high-frequency stress wave that will travel through the material. There is, therefore, great benefit from a health monitoring perspective in being able to spatially locate these signals, particularly for larger structures, providing the platform for more informed maintenance strategies. 

A fundamental component in many of these data-driven approaches is learning acoustic emission difference-in-time-of arrival ($\Delta T$) mappings \cite{jones2020bayesian,baxter2007delta,pearson2017improved,ebrahimkhanlou2018single}, which spatially characterise the difference in arrival time between two sensors of an AE source across the surface of a structure. Although promising, this method is severely hampered by the requirement to collect an extensive set of artificial sources over the entire structure, making the action of constructing the maps costly from both a financial and temporal perspective. This cost is particularly pertinent as the size of the structures considered grows. It is, therefore, critical that the size of the training set is minimised. The challenge here is then scaling to structures that contain complex geometrical features such as holes and bolts, which will disrupt the propagation path of the ultrasonic waves. These irregularities will induce sharp discontinuities in the AE arrival time, and so a standard data-driven learner will require a dense training set in these regions to suitably capture the feature behaviour, conflicting with the desire to reduce the amount of data that needs to be acquired. An additional challenge is faced from the action of retrofitting a monitoring system, where gaining access to the complete coverage of all parts of a structure is often challenging, particularly at locations that contain joints and interfaces. As such, it is important to obtain good model generalisation in areas where measurements may be more sparse, which proves difficult when solely reliant on a data-driven learner. These problems are not unique to the application of acoustic emission localisation however, and instead arise consistently wherever black-box learners are deployed. Whilst their flexible nature enable a high level of performance in the presence of an abundance of data, when presented with sparse training sets or forced to extrapolate, there is often no guarantee on how the predictions will behave. In this work, we propose a means of addressing the above limitations of methods that rely on $\Delta T$ mapping by incorporating physical knowledge into the learning process. Methodologies that consider such an approach can broadly be grouped under the term physics-informed machine learning, where physical insight is embedded into data-driven algorithms - for a general overview of this field, see \cite{willard2020integrating}. Also referred to as grey-box modelling \cite{cross2022physics}, the objective when adopting such a model architecture is to combine the expressive power of machine learning tools with physically derived laws or constraints. Predictions that are drawn from these models are then guaranteed to be representative of some underlying physical laws that govern the dynamics of the system under consideration.   

%The problems outlined above are not unique to the application of acoustic emission localisation however, and instead arise consistently wherever black-box learners are deployed. Whilst their flexible nature enable a high level of performance in the presence of an abundance of data, when presented with sparse training sets or forced to extrapolate, there is often no guarantee on how the predictions will behave. For example, \cite{cross2022physics} investigates strain prediction on an aircraft wing during flight. When the model is tasked with prediction across flight conditions not seen in the training phase, it can be seen that the model is unable to generalise and returns less accurate strain predictions than where strain is predicted under flight conditions similar to those used in training. 

One particular way to embed physically-derived insight is by constraining the learning algorithm with physical constraints such that subsequent predictions comply with these assumptions. In the context of Gaussian process regression \cite{williams2006gaussian}, a Bayesian machine learning tool often employed in SHM, there are numerous ways constraints can be applied, as extensively discussed in \cite{swiler2020survey}. For example, if one has knowledge of the shape of the latent function, then monotonicity or convexity constraints can be applied \cite{agrell2019gaussian,riihimaki2010gaussian,da2012gaussian,maatouk2017finite,lopez2018finite}. General bounds such as nonnegativity can also be imposed on the GP prior \cite{pensoneault2020nonnegativity}, as well as constrained to satisfy linear operators \cite{jidling2017linearly}. Where more specific insight is available, for example, the underlying equation of motion, derivation of an exact autocovariance is possible \cite{cross2021physics}. Constraints also exist in the form of vector-output Gaussian processes, where relevant physical relationships between multivariate targets can be embedded into the cross-covariance terms. Under a multivariate output framework, the inclusion of both ordinary differential equations \cite{alvarez2009latent,raissi2018hidden} and boundary conditions have been explored \cite{cross2019grey}. 

%If the interest is in extending to the latter stage of the SHM paradigm, the full range of operating conditions required to be considered extends to include all possible damage states. Given that engineering assets are often of high-value, the cost associated with such a procedure clearly makes this action impractical for most scenarios. 

%Multi-output Gaussian processes will not be discussed further in this paper, and so for proper treatment of the topic, the reader is referred to a number of helpful references.

%constrained Gaussian processes that enable one to bound predictions to a fixed spatial domain will be explored in the context of learning acoustic emission (AE) difference-in-onset time $(\Delta T)$ maps. 

The focus of the work presented here will be on the application of constrained Gaussian processes for learning difference-in-time of arrival/$\Delta T$ acoustic emission maps. The nature of the constraints that are considered are those of physical boundary conditions, embedded into the model by firstly rewriting the covariance of the GP prior as a Laplacian eigenfunction expansion. Given that the eigenfunctions are unique to a user-specified domain, the GP can then be naturally constrained to some boundary conditions along a physical domain. This results in a model that retains the flexibility of a traditional machine learner, whilst adhering to known physical conditions that exist at boundary locations. To the authors' best knowledge, up until now, all previous works that consider a data-driven approach to AE localisation have been purely black-box in nature. This paper demonstrates that by ensuring that the model is constrained to known physical laws, the process of generating $\Delta T$ maps is made feasible, improving the predictive performance in cases of sparse/few training observations and where training measurements only partially cover the full structure spatially, and therefore, only partial coverage of the input space for the data-driven learner. General discussions surrounding where one may implement the constrained Gaussian process are also considered. 

It should be noted that where suitably dense training data is available across the whole structure, it is not expected that the constrained GP will significantly improve performance, particularly where training data is available on boundary locations. In the case where the training set includes observations on and around a boundary, then these points can be seen seen as constraints themselves in the sense that the resulting predictions will be constrained to these measurements. The purpose of this paper, however, is to explore how adding physical-consistency into the GP prior can assist where acquiring data is challenging and thus training data is limited, as often occurs in the $\Delta T$ mapping procedure. 

The paper proceeds as follows; Section 2 offers a brief introduction to Gaussian process regression, and then outlines the necessary theory for constraining a GP. Section 3 details the general procedure of constructing $\Delta T$ maps, including details of the data set used throughout the work. A practical discussion of how one can implement constraints in this context is then given. Section 4 presents the results and corresponding discussion, with concluding remarks offered in Section 5.

%We show that through constraining a Gaussian process by boundary conditions, predictive performance can be improved in scenarios where data coverage of the input space is restricted, such as when access to the entire physical structure is limited, as well as when training data becomes sparse.

%where sharp discontinuities are often induced in the AE features that can be challenging to learn with a pure machine learner without a dense training set *check this based on 1D plots*

\section{Constrained Gaussian processes}

%By virtue of their flexible nature, it can be shown that for common kernel choices, they are universal approximators in the sense that they can approximate any continuous function provided sufficient training data \cite{micchelli2006universal}.

Gaussian processes provide a Bayesian, non-parametric tool for solving regression problems. In a regression context, a Gaussian process specifies a prior distribution over a latent function, $f(\mathbf{x}) : \mathbf{x} \in \mathbb{R}^d$, which in addition to a noise term $\epsilon$, is believed to represent the target value $y$ such that,

\begin{equation}\label{eq:2.1}
    y = f(\mathbf{x}) + \epsilon \hspace{5mm} \epsilon \sim \mathcal{N}(0,\sigma_n^2),
\end{equation}

\noindent where,

\begin{equation}\label{eq:2.2}
    f \sim GP(m(\mathbf{x}),k(\mathbf{x},\mathbf{x}'))
\end{equation}

\noindent It can be seen in the above equation that a Gaussian process is fully defined by a mean function, $m(.)$ and a covariance function $k(.)$, which together, characterise one's prior belief about the behaviour of the latent function. The mean function can be specified as any basis function expansion of $\mathbf{x}$, whilst also having the ability to generalise to an input space $\mathbf{x}_m \in \mathbb{R}^q$, where $q$ can differ from the input space dimension $d$. The covariance function then captures the covariance between two points in the input space.

As a set of training data $\{X,\mathbf{y}\}$ become available, it is possible to condition the Gaussian process prior on this set of known input/outputs to form a posterior $\mathrm{y}_\star$, over an unknown input, $\mathbf{x}_\star$. Following \cite{williams2006gaussian}, simple Gaussian machinery allows us to arrive at a closed form expression for the distribution over $y_\star$,

\begin{equation}\label{eq:2.3}
    p(\mathrm{y}_\star | \mathbf{x}_\star, X,\mathbf{y}, \mathbf{\theta}) = \mathcal{N}(\mathbb{E}[\mathrm{y}_\star],\mathbb{V}[\mathrm{y}_\star]),
\end{equation}

\noindent where, 

\begin{align}\label{eq:2.4}
    \mathbb{E}[\mathrm{y}_\star]& = m(\mathbf{x}_\star) + K(\mathbf{x}_\star,X)(K(X,X)+\sigma_N^2\mathbb{I})^{-1}\mathbf{y}, \\
    \mathbb{V}[\mathrm{y}_\star]&= K(\mathbf{x}_\star,\mathbf{x}_\star) - K(\mathbf{x}_\star,X)(K(X,X)+\sigma_N^2\mathbb{I})^{-1}K(X,\mathbf{x}_\star)). \label{eq:2.5}
\end{align}

\noindent When specifying a covariance, there will often be a number of hyperparameters $\mathbf{\theta}$ that are required to be specified (note that $\mathbf{\theta}$ can also incorporate coefficients of a mean parametric function). These hyperparameters will alter the behaviour of the chosen kernel and often have a meaningful interpretation. For example, many popular kernels contain a length scale parameter, which in a practical sense, governs how close inputs in the same dimension should be to influence one another. Due to the framework in which the Gaussian process resides, it is possible to determine the value of these parameters in a systematic manner by optimising the marginal likelihood of the model (also known as the model evidence). Should the reader be interested, information detailing this procedure is included in \ref{app:appendix1}.

%include "Bayesian framework in which the Gaussian process resides" in thesis

\subsection{Embedding physical constraints into a Gaussian process}

Although Gaussian processes present a powerful tool, as with all machine learners, they are still black-box in the sense that their performance is entirely reliant on the data that the model is trained on. In cases where sufficient training data are scarce, then the resulting model may struggle to adequately learn the underlying behaviour of the features. As introduced in Section 1, one way to circumvent such issues is to incorporate physical insight into the machine learner. In this paper, focus is directed on embedding boundary condition knowledge into a Gaussian process prior through constraining the covariance function. This presents one view of a constrained process. From the perspective of the nature of the physics considered, the advantage here is that the level of insight into the governing mechanistic laws can be relatively shallow; knowledge of boundary conditions are generally easier to come by than an exact governing differential equation. Additionally, through directly constraining the form of the prior, the approach does not rely on the addition of artificial observations at the boundary. In fact, the method employed here is a sparse approximation, and so is computationally cheaper than the standard implementation of a Gaussian process, both in terms of complexity and storage demands \cite{solin2020hilbert}.

To constrain a Gaussian process in this manner, it is first necessary to make use of the following covariance function approximation \cite{solin2020hilbert}: 

\begin{equation}\label{eq:2.6}
    k(\mathbf{x},\mathbf{x}') \approx \sum_j^m S(\sqrt{\lambda_j})\phi_j(\mathbf{x})\phi_j(\mathbf{x}')
\end{equation}

\noindent Under this representation, the covariance function is defined as a basis function expansion across $m$ Laplacian eigenfunctions $\phi$ of a user-selected domain, projected onto the spectral density $S$ of the covariance that has been evaluated in a point-wise manner at the corresponding Laplacian eigenvalues $\lambda$. To calculate the Laplacian eigenpairs, one is required to solve an eigenvalue problem of the form,

\begin{equation}\label{eq:2.7}
    -\nabla^2\phi_{j}(\mathbf{x}) = \lambda_{j}^2\phi_{j}(\mathbf{x}), \hspace{5mm} \mathbf{x} \in \Omega,
\end{equation}

\noindent where $\nabla^2$ is the Laplacian operator and $\Omega$ represents the domain of interest. As for any differential equation, its solution may be sought given boundary conditions described generally by,

\begin{equation}\label{eq:2.7a}
    \Psi[\phi_{j}(\mathbf{x})] = H(\mathbf{x}),  \hspace{5mm} \mathbf{x} \in \delta \Omega,
\end{equation}

\noindent where $\Psi$ denotes some operator and $H$ is an arbitrary function that maps $\mathbf{x}$ to a known value which exists on the boundary of a domain $\delta \Omega$. The solutions for $\lambda$ and $\phi$ are, therefore, bound to the chosen domain $\Omega$, and, consequently, are unique to the boundary conditions specified in equation (\ref{eq:2.7a}). Upon substitution into equation (\ref{eq:2.6}), each draw from the prior is then guaranteed to abide by these constraints, returning an expression for the covariance that is dependent on the boundary conditions of the feature space.

To specify a suitable kernel spectral density, it is possible to employ Bochner's theorem which states that the covariance of a stationary function can be represented by the Fourier transform of a positive, finite measure \cite{stein2012interpolation}. If this measure has a corresponding density $S$, then the spectral density and the covariance are Fourier duals of one another \cite{chatfield2019analysis}. For example, the spectral density of the Mat\'ern 3/2 kernel takes the following functional form,

\begin{equation}\label{eq:2.9}
    S(\omega) = \sigma_f^2 \frac{4\pi (2\nu)^\nu}{l^{2\nu}} \frac{\Gamma(\nu+1)}{\Gamma(\nu)}(\frac{2\nu}{l^2}+w^2)^{(-\nu+1)},
    \end{equation}
    
\noindent where $\nu=3/2$ and $\Gamma$ denotes the Gamma function. For a test point $\mathbf{x}_\star$, the predictive posterior is then defined as:

\begin{align}\label{eq:2.10}
    \mathbb{E}[\mathrm{y}_{\star}] &= \boldsymbol{\Phi}_{\star}(\boldsymbol{\Phi}'\boldsymbol{\Phi}+\sigma_n^2\boldsymbol{\Lambda}^{-1})^{-1}\boldsymbol{\Phi}'\mathbf{y} \\
    \mathbb{V}[\mathrm{y}_{\star}] &=  \sigma_n^2\boldsymbol{\Phi}_{\star}(\boldsymbol{\Phi}'\boldsymbol{\Phi}+\sigma_n^2 \mathbf{\Lambda^{-1}})^{-1} \boldsymbol{\Phi}_{\star}',\label{eq:2.11}
    \end{align}

\noindent where,

\begin{align}\label{eq:2.12}
    \boldsymbol{\Phi} &= (\phi_{1}(X),\phi_{2}(X),\phi_{3}(X),...,\phi_{m}(X)), \\
    \boldsymbol{\Lambda} &= \diag(S(\lambda_{1}),S(\lambda_{2}),S(\lambda_{3}),...,S(\lambda_{m})),\label{eq:2.13}
\end{align}

 \noindent with the mean function now set to zero. The reasoning for this is not due to the mean being uninteresting; that's quite the opposite in a grey-box context. However, specifying a physics-based basis function often requires a significant level of knowledge regarding the dynamics of the system being modelled. The intention here is to present a general method that can be applied with known boundary condition information, which, in most cases, is readily available. For interesting examples of applying a physics-based mean function, the reader is referred to the following works \cite{pitchforth2021grey,zhang2020gaussian}.  

 Finally, as shown in equation (\ref{eq:2.9}), there are a number of hyperparameters to learn. Again, a type-II maximum likelihood approach as detailed in \ref{app:appendix1} can be followed.

%\begin{equation}
%    \mathbf{f} \sim \mathcal{GP}
%\end{equation}

\section{Physical constraints for acoustic emission time of flight mapping}

% make title more general? acoustic emission replaced with ultrasonic or perhaps completely removed? application of ** to acoustic emission time of flight mapping?

This section will introduce acoustic emission difference-in-time of arrival mapping, and how physically-relevant constraints may be incorporated into models that utilise such features. The data set used throughout the paper will then be detailed, which consists of $\Delta T$ measurements from a plate structure. The structure itself presents a challenging wavefield to model, containing a series of holes cut through the plate that adds significant complexity to the wave propagation behaviour. The section then concludes with a discussion on how constraints can be implemented.

\subsection{Acoustic emission onset time mapping}

Acoustic emissions characterise ultrasonic signals that are released as a structure undergoes some internal change. Often these changes are initiated by mechanisms such as crack propagation, spalling and delamination, making acoustic emission measurements highly suitable for use as features in damage monitoring strategies. Given that the time taken for an AE signal to arrive at a receiving sensor will be dependent on the distance traveled, it is possible to use acoustic emission measurements as a means for localisation \cite{tobias1976acoustic,kundu2014acoustic}. In particular, methods that view AE localisation as a problem of spatial mapping of $\Delta T$ information have been proven to perform well in challenging localisation environments \cite{jones2020bayesian,baxter2007delta,ebrahimkhanlou2018single}. There, some regression algorithm is first employed to learn the spatial variation in $\Delta T$ across a structure of interest from a set of training measurements. An estimate for the source origin of a future AE event is then inferred as the location that minimises the difference between the observed and predicted $\Delta T$ \cite{baxter2007delta,al2016acoustic}, or by considering the source location likelihood as part of a probabilistic framework providing that the spatial map has been learnt as a distribution \cite{jones2020bayesian}. In the latter case, not only is a maximum likelihood estimate of the source location provided, but also an associated confidence, which may be used to inform a maintenance engineer how large of an area to inspect, for example, in a wind turbine bearing \cite{jones2021bayesian}.

\subsection{Experimental case study}

\begin{figure}[h]\label{fig:f1}
    \centering
    \subfloat[Schematic of plate with sensor positions labeled]{\includegraphics[scale=0.5]{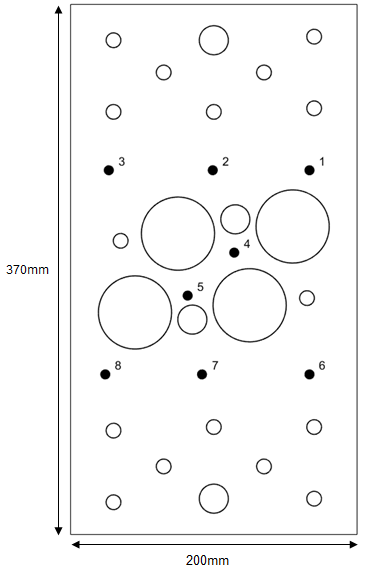}\label{fig:f1a}}
    \hspace{10mm}
    \subfloat[Real image of plate]{\includegraphics[scale=0.5]{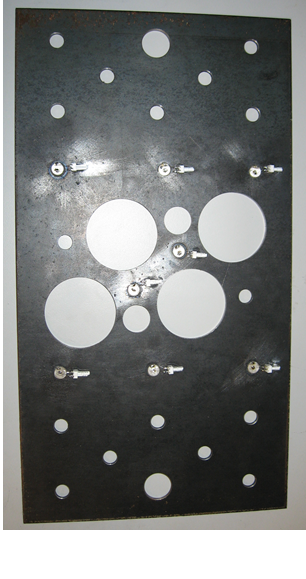}\label{fig:f1b}}
    \caption{Plate structure used as a case study. Recreated from \cite{hensman2010locating}.}
  \end{figure}

In this work, we will consider learning $\Delta T$ maps for a complex plate structure. First used in the work of
Hensman et al. \cite{hensman2010locating}, the plate was manufactured to contain a series of holes to replicate the challenges found in real
engineering infrastructure. These holes, which can be seen on the schematic of
the plate in Figure \ref{fig:f1a}, induce a number of complex phenomena such as
scattering and wave mode conversion. For large areas of the plate, a direct
propagation path for many of the sensor pairs is also blocked, adding further
complexity.

\begin{figure}[h]
    \centering
    \includegraphics[scale=0.98]{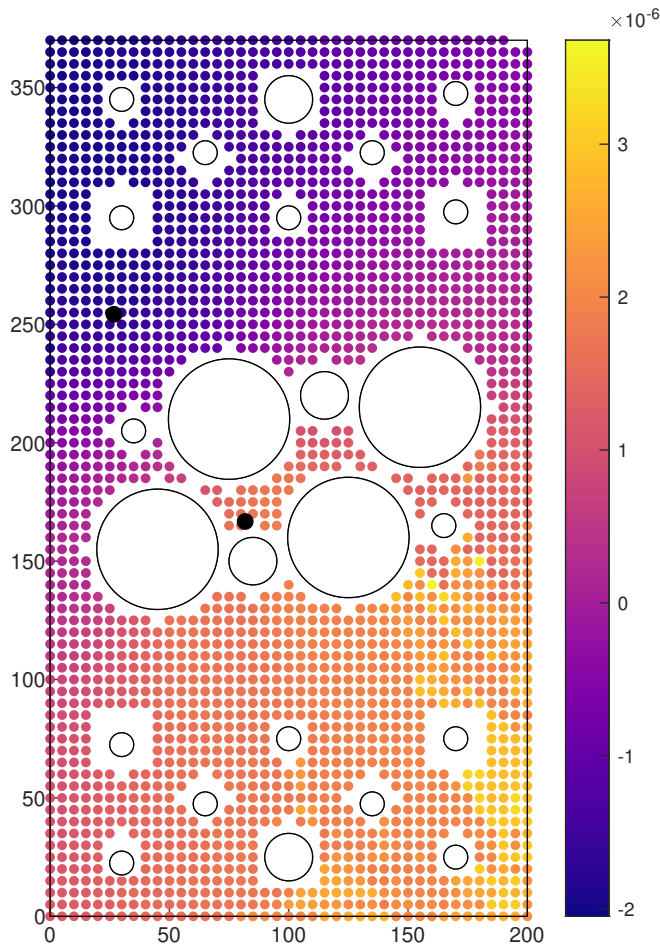}
    \caption{True difference-in-time-of arrival ($\Delta T$) values for sensor pair 3 \& 5 across all of the positions on the plate that data was acquired. Sensor locations highlighted by the black circles.} 
    \label{fig:f3.2}
\end{figure}

The dataset acquired in \cite{hensman2010locating} is used for the entirety of this work. Should the reader be interested in a complete description of the experimental and data acquisition procedure, they should consult the original paper. However, in brief, artificial AE events were first generated in a uniform grid across the surface of the plate. The time of arrival of each event was then captured at every sensor, returning an eight dimensional onset time vector for each artificial excitation. Given that there are 28 possible pair-wise combinations, it is then trivial to transform this vector into a 28 dimensional vector of $\Delta T$ values. As an example, the full set of true $\Delta T$ values, which form the targets of the GP, for sensor pair 3 \& 5 is shown in Figure \ref{fig:f3.2}, consisting of 2277 measurement locations, each spaced 5mm apart except for where holes in the plate are present.

\subsection{Implementing constraints}

%To enhance. In particular, to allow so that b, TO overcome. A primary reason for selecting dTOA mapping as a case study is that it provides a practical engineering problem for which spatial boundary conditions naturally exist in. Further to this, the plate structure is particularly relevant since there is ample opportunity for embedding constraints, with boundaries at both the limits of the domain in addition to the many holes located across the plate. 

% maybe add back in - From the perspective of a dynamicist, although not representing the same physical phenomena, the mathematical , the this process can be seen as analogous to finding the natural frequency and mode shapes of the plate. 

To implement the constrained GP, it is firstly required that a physical domain is specified, and then for the associated Laplacian eigenfunctions to be solved. As these eigenfunctions are unique to a given domain, it is possible to directly encode boundary condition knowledge into the model. As we are using a Laplacian approximation, from the perspective of a dynamicist, this process can be seen as analogous to finding the wavenumbers and approximate mode shapes of the plate. For simple geometries, it is possible to arrive at closed form solutions for the Laplacian eigenvalues. However, due to the geometrical complexity of the plate, in this work the eigendecomposition is computed numerically. This is calculated by approximating the operator with a finite difference equation that is solved alongside the boundary conditions, with each boundary condition giving an equation that can be solved simultaneously with equation (\ref{eq:2.7}). To implement this numerical approximation, the Laplacian operator is first converted into its discrete counterpart by transforming the domain into a \textit{grid mask}, $u$, which exists as a binary matrix where ones denote locations inside the domain, whilst zeros indicate the opposite. A discrete representation of the Laplacian can then be formed as a stencil matrix by applying a finite difference approximation of the Laplace operation on $u$, 

\begin{equation}\label{eq:3.1}
    -\nabla^2 u(i,j) \approx \frac{1}{h^2}(-4u_{i,j} +u_{i-1,j} + u_{i,j-1} + u_{i+1,j} + u_{i,j+1}),
 \end{equation}

\noindent where $(i,j)$ index the rows and columns of the grid mask, and $h$ represents the step size between adjacent nodes within the grid. Equation (\ref{eq:3.1}) can then be manipulated to reflect known information solution information in the form of boundaries conditions where $(i,j)$ lie on the boundary of the grid mask. For example, in the case of Dirichlet conditions (process value is specified), the solution can be fixed at the boundary positions. For the onset time functions, the associated boundary condition is that of a first order spatial derivative equal to zero (Neumann boundary conditions). Equation (\ref{eq:2.7a}) can, therefore, be rewritten as

\begin{equation}\label{eq:3.2}
    d \phi_{j}(\mathbf{x}) = 0  \hspace{5mm}  \mathbf{x} \in \delta \Omega.
\end{equation}

\begin{figure}[h]
    \centering
    \includegraphics[scale=0.575]{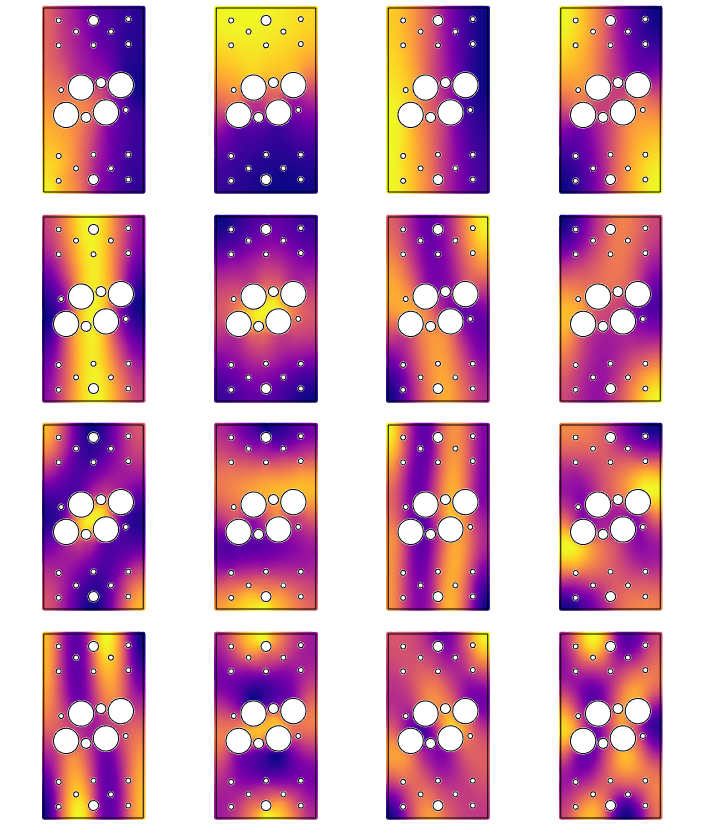}
    \caption{First 16 Laplacian eigenfunctions (left to right) of the plate detailed in Section 3.2} \label{fig:3.3}
\end{figure}

\noindent Following the construction of a numerical approximation of the negative Laplacian of $\Omega$, for which the procedure of solving for first order boundary conditions is detailed in \ref{appendix2}, the leading $m$ eigenvalues and eigenfunctions can then be calculated through a chosen numerical solver. For the work conducted in this paper, following \cite{solin2019know} for large-scale spatial mapping problems, 256 eigenbases are used. The first 16 of these basis functions that incorporate all physical boundaries present on the plate are shown in Figure \ref{fig:3.3}. Note that although we only consider homogenous first order derivative boundaries, it is possible to include other forms of boundary condition through proper treatment of equation (\ref{eq:3.1}) at boundary locations.

\section{Results and discussion}\label{sec:4}

%In the following section, the performance of the constrained GP for AE onset mapping is considered across a number of case studies. The objective is to examine what scenarios that an SHM practitioner would want to consider employing a constrained model over a standard Gaussian process or arbitrary machine learner, and where it will prove useful both in the context of AE localisation and spatial mapping generally. We begin by examining each model as training data becomes more sparse, particularly as the density and availability of measurements at the boundary of the domain varies. Results and discussion will then be presented from the perspective of when there is only partial coverage of the structure available, and therefore measurements are not available across the entirety of the domain. In all cases explored, the test set remains consistent, containing 2277 measurements collected at a 5mm uniform spacing across the whole plate.

The feasibility of acoustic emission localisation for large and complex systems is severely hampered by the need to collect artificial AE events at locations on a dense grid across the structure. To explore how the constrained GP may help to alleviate this burden, in this section, results are shown that investigate scenarios where the number and location of training points are limited. To mimic the likely availability of training data from a measurement campaign on a real structure, we consider firstly the case where training measurements are available across the structure but with limited grid density. For most structures, however, particularly those with many connecting components, it is unlikely that access to the whole structure would be available to establish a training dataset (e.g.\ where access is obscured, or between closely spaced components). The second scenario investigated, therefore, limits the training dataset to a single part of the plate.

The investigation will compare the performance of the standard and constrained GPs. Naturally, the availability of measurements themselves from the boundary for GP conditioning will affect the performance of both methods. We quantify this by explicitly considering additional measurements at the available boundary for both scenarios. In the first case, where measurements are available across the full structure, we expect that the standard GP will outperform the constrained model when training grids are dense - the constrained GP is after all an approximation. We expect to see the benefit of the constrained model where training data are sparse and when data are only available from limited locations across the structure (as will likely be the case in operation). 

%As a measure of model predictive performance, the normalised mean square error (nMSE) and mean standardised log loss (MSLL) are considered across a test set of all 2277 data points, collected at a uniform spacing of 5mm across the whole plate. The nMSE is defined as

As a measure of model predictive performance, the normalised mean square error (nMSE) is considered across a test set of all 2277 data points, collected at a uniform spacing of 5mm across the whole plate. The nMSE is defined as

\begin{equation}
    nMSE = \frac{100}{N\sigma_y^2}(\mathbf{y}_\star-\mathbf{y}^*)^T(\mathbf{y}_\star-\mathbf{y}^*),
\end{equation}

\noindent where $N$ is the number of points, $\sigma_y^2$ is the variance of the true targets, $\mathbf{y}_\star$ are the model predictions, with $\mathbf{y}^*$ the true targets. A score of 0 would be returned if the model predictions perfectly aligned with the true targets, whilst a score of 100 is identical to predicting at the mean.

\subsection{Sparse training data available across the whole structure}

When undergoing an AE data collection campaign, it is often not practical to collect a dense grid of training observations, particularly for larger structures or when setting up multiple monitoring regions. To consider how the constrained GP performs where the training data availability is reduced, a number of training sets containing varying numbers of observations are formed, with measurements available across the full spatial limit of the structure. For an individual set, the spacing between training points is uniform (excluding where the holes are located), with the value that the spacing is fixed at varying across the sets. The total range of training sets considered is outlined in Table \ref{table:4.2}. 

\begin{figure}[h]
    \includegraphics{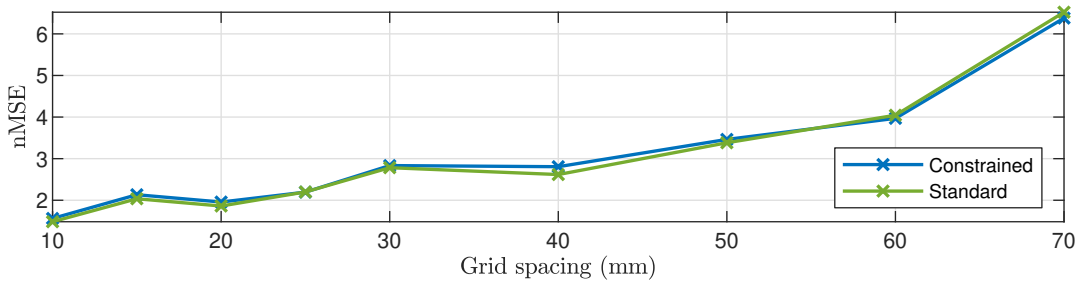}%
    \caption{nMSE of the predictions made by the constrained and standard Gaussian process as the number of training points are reduced. Here the training points fully cover the plate, but at different grid densities.}
    \label{fig:4.1a}
\end{figure}

The nMSE returned on the test set for each training set, averaged over all available sensor pairs, is plotted on Figure \ref{fig:4.1a} respectively for both the constrained and standard GP. 

\begin{table}[h]
    \centering
    \begin{tabular}{llllllllll}
    \hline
    \multicolumn{1}{l|}{Spacing between training points (mm)} & 10  & 15  & 20  & 25  & 30 & 40 & 50 & 60 & 70 \\
    \multicolumn{1}{l|}{Number of observations in training set}         & 600 & 302 & 160 & 107 & 71 & 50 & 32 & 14 & 11 \\ \hline  
    \end{tabular}
    \caption{Spacing between observations and corresponding total number of datum points for each training set used.}
    \label{table:4.2}
\end{table}

The figure demonstrates that as the number of training points reduced, the constrained GP offers a greater nMSE accuracy. At the lower end of the grid spacings where the training set is denser, it can be seen that the standard GP is slightly favourable, which as introduced above, is expected. Where training data coverage and availability is good, sufficient boundary condition insight can be obtained from training measurements. That is not to say, however, that one should always seek to retain the full covariance where data are easily assessable. When computing predictions with the constrained GP, computational complexity reduces from $\mathcal{O}(n^3)$ to $\mathcal{O}(nm^2)$, with storage requirements moving from $\mathcal{O}(n^2)$ to $\mathcal{O}(nm)$. A benefit here, therefore, is that when the number of training points exceed several thousand, the use of the constrained GP presents a practical solution \cite{jones2020isma} without turning to more complex computing techniques such as parallelisation and/or the use of graphical processing units.   

% \begin{figure}[htp]

%     \subfloat[nMSE\label{fig:4.1ai}]{%
%       \includegraphics{images/alpha_A_nmse.eps}%
%     }

%     \subfloat[MSLL\label{fig:4.1aii}]{%
%       \includegraphics{images/alpha_A_msll.eps}%
%     }
    
%     \caption{nMSE and MSLL of the predictions made by the constrained and standard Gaussian process as the number of training points are reduced. Here the training points fully cover the plate, but at different grid densities.}
%     \label{fig:4.1a}
% \end{figure}

\begin{figure}[h]
    \includegraphics{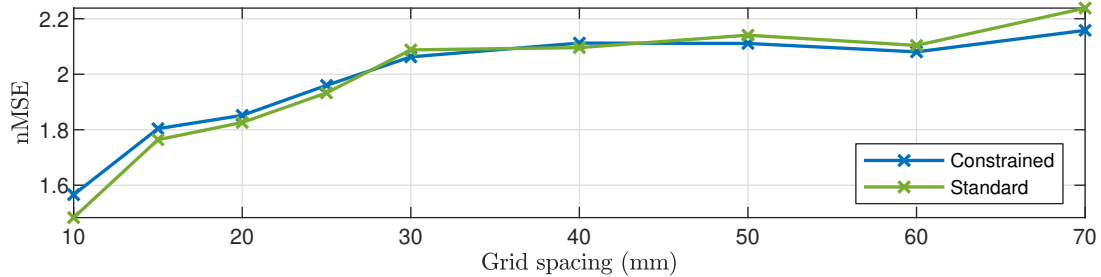}%
     \caption{nMSE of the predictions made by the constrained and standard Gaussian process as training points are reduced whilst retaining full boundary measurements.}
    \label{fig:4.1b}
\end{figure}

In the above example, it was assumed that boundary measurements were available inline with the overall training grid density. For example, at a spacing of 20mm, boundary measurements were available every 20mm. It may arise, however, that one would wish to gather more insight in regions on or around boundaries. For example, when mapping an acoustic emission wavefield, it is likely that one would want more insight around boundaries of the domain, where sharp discontinuities will be introduced into the propagation pattern. The first option available in this scenario would be to collect more measurements at the boundary location, which we consider by repeating the above experiment, but ensure that boundary measurements are available every 10mm\footnote{For the outside boundaries, measurements were taken directly on the boundary. For the inside boundaries, measurements were not available directly on the boundary. Therefore, inside boundary measurements were defined as data points directly adjacent to a given hole in line with a 10mm grid spacing.}. This results in a scenario where one may take a fairly sparse grid of training measurements, but adopt a more fine resolution at boundary locations. Clearly, it is also possible to combine more training measurements with physical constraints. Figure \ref{fig:4.1b} plots the results for both of these cases. 

% \begin{figure}[htp]

%     \subfloat[nMSE\label{fig:4.1bi}]{%
%       \includegraphics{images/alpha_B_nmse.eps}%
%     }

%     \subfloat[MSLL\label{fig:4.1bii}]{%
%       \includegraphics{images/alpha_B_msll.eps}%
%     }
    
%     \caption{nMSE and MSLL of the predictions made by the constrained and standard Gaussian process as training points are reduced whilst retaining full boundary measurements.}
%     \label{fig:4.1b}
% \end{figure}

% One sentence talking about msll removed - 4 lines down

\begin{figure}[h]
    \includegraphics{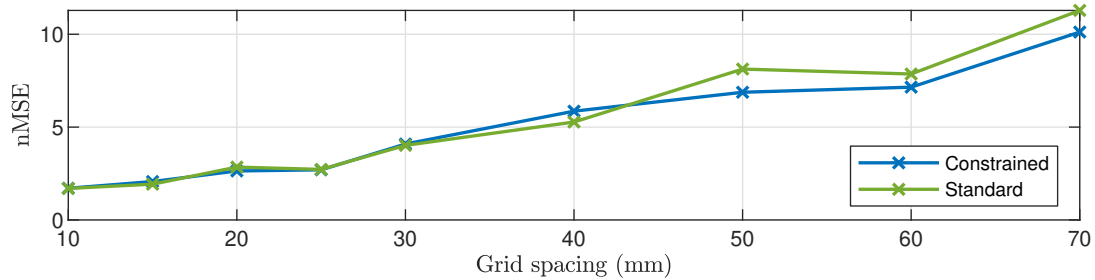}
    \caption{nMSE of the predictions made by the constrained and standard Gaussian process as training points are reduced with no boundary measurements.}
     \label{fig:4.1c}
\end{figure}

It can be seen that both models obtain similar error scores, with a small improvement returned by the constraints where training data is very sparse. In the case that a measurement campaign has been specifically conducted with both extra measurements at the boundary and a dense training grid, then the benefit of using the constrained GP from the perspective of mean predictive accuracy is negligible. There are, however, a number of disadvantages to simply adding more training points at the boundary. As the structures we wish to represent become more complex, particularly when in two or three dimensions, the number of datum points required to sufficiently capture a continuous domain will quickly grow. Given the cubic and quadratic scaling of complexity and storage respectively for standard GPs, prohibitive computational demands can quickly be reached, and so for large or complex structures that will demand many boundary measurements, the constrained GP will be a more feasible solution. An additional limitation when collecting data is that many engineering structures simply prevent acquisition at boundaries, for example, at joints and connections that physically obscure generating an artificial signal at that location. We, therefore, now examine a second scenario, where one has access to no boundary measurements, repeating the procedure in the preceding paragraph, but removing all boundary locations from the training set. The predictive performance on the test set for both forms of model is plotted in Figure \ref{fig:4.1c}. The figure demonstrates an improved performance from the constrained GP as training data become fewer. The significance here however, is that performance gain of the constrained GP at larger grid spacings is higher than that obtained in Figure \ref{fig:4.1a} and \ref{fig:4.1b}, where the error returned by both models is comparable for the denser training sets. As the standard GP now has no knowledge of boundary conditions, it is forced to predict at boundary locations with little information if the training grid is not dense. In the case of the constrained GP, the constraints provide the kernel function additional information to the training measurements, incorporating physically relevant structure into the covariance that can then be used when predicting at locations on and adjacent to boundary locations.

% \begin{figure}[htp]

%     \subfloat[nMSE\label{fig:4.1bbi}]{%
%       \includegraphics{images/alpha_C_nmse.eps}%
%     }

%     \subfloat[MSLL\label{fig:4.1bbii}]{%
%       \includegraphics{images/alpha_C_msll.eps}%
%     }
    
%     \caption{nMSE and MSLL of the predictions made by the constrained and standard Gaussian process as training points are reduced with no boundary measurements.}
%     \label{fig:4.1bb}
% \end{figure}#

Overall, what can be deduced from the results in this section is that when fewer training points are available, the inclusion of boundary constraints improve the predictive capabilities of the model, particularly where boundary measurements are sparse or unavailable.  Whilst there is no guarantee that an improved predictive accuracy will be obtained where training data are in more of an abundance, particularly at the boundaries, the use of the constrained GP will still be a consideration when one is limited by the computational demands of calculating the predictive equations in closed form through the standard GP implementation. 

\subsection{Training data from partial structure coverage}

When undergoing an SHM data collection campaign, as previously discussed, it is often not possible to collect data across the entire input space. In a spatial mapping context, this limitation generally arises through being unable to collect data that fully covers the entire structure of interest. For example, when acquiring the artificial AE events that are used to learn a $\Delta T$ mapping, it may not be possible to gain full access of the structure, particularly when a health monitoring system is being retrofitted. For example, such a scenario may arise if trying to collect data from the drive train of a wind turbine gearbox, where the assembly of various interlocking gears and shafts will obscure access to many of the individual components that one may be interested in developing an AE mapping for. The fuselage of an aircraft is another example where full access is prevented without disassembly, particularly in the areas where the root of the wings are mounted. 

\begin{figure}[h]
    \includegraphics{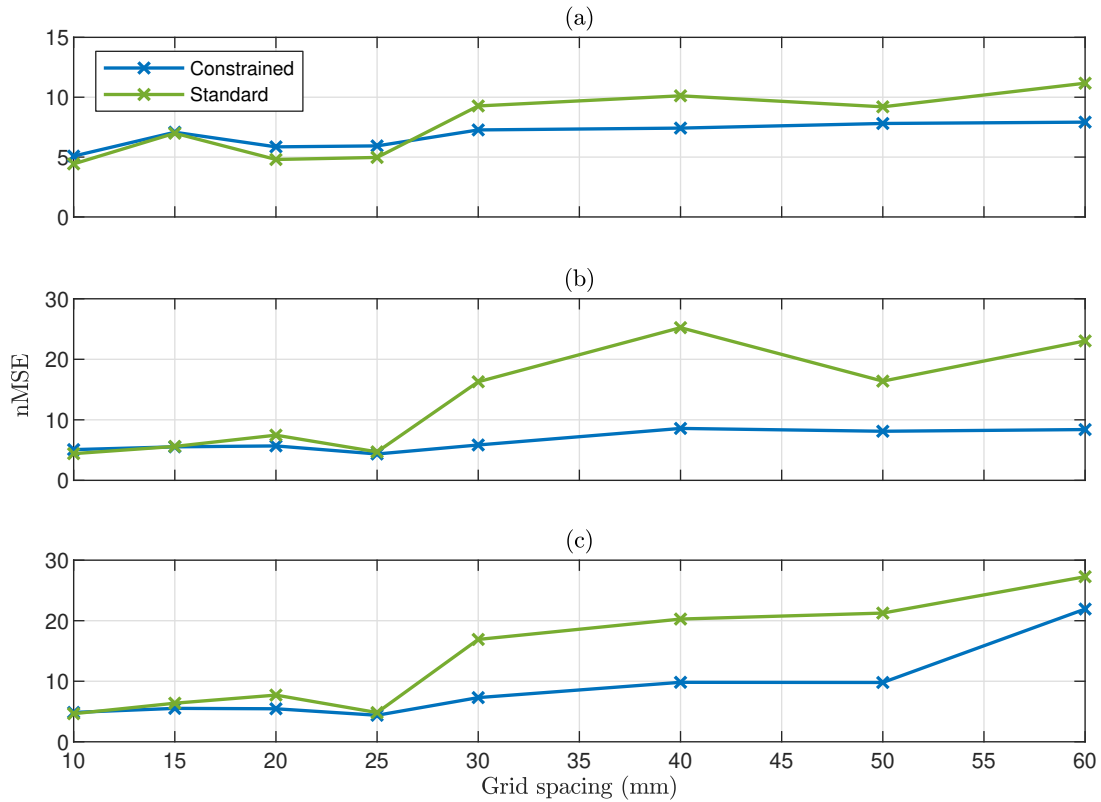}%
    \caption{Predictive performance with restricted training coverage for a) full boundary location measurements b) partial boundary measurements c) no boundary measurements.}
    \label{fig:4.1d}
\end{figure}

To explore how the constrained GP can mitigate against a lack of training data coverage, the data points used to train the models are now restricted to the middle section of the plate. The three training conditions with respect to the inclusion of boundary points in the training set are again considered; these are a) full boundary coverage, b) partial boundary coverage (inline with the overall training grid density), and c) no boundary measurements. Predictive performance on the test set for each of these three conditions for both the constrained and standard GP is shown in Figure \ref{fig:4.1d}. 

% \begin{figure}[htp]

%     \subfloat[nMSE]{%
%       \includegraphics{images/beta_A_B_C_nmse.eps}%
%     }
    
%     \subfloat[MSLL]{%
%       \includegraphics{images/beta_A_B_C_msll.eps}%
%     }
    
%     \caption{Predictive performance with restricted training coverage for a) full boundary location measurements b) partial boundary measurements c) no boundary measurements.}
%     \label{fig:4.1c}
% \end{figure}

Across all three cases, the constrained GP offers either an improved or comparable predictive performance at all of the training point spacings. For the increased grid spacings, particularly where partial or no boundary measurements are available, a large performance gain is obtained by constraining the GP, regularly exceeding an nMSE reduction of greater than 10\%. 

\begin{figure}[h]
    \centering
    \includegraphics{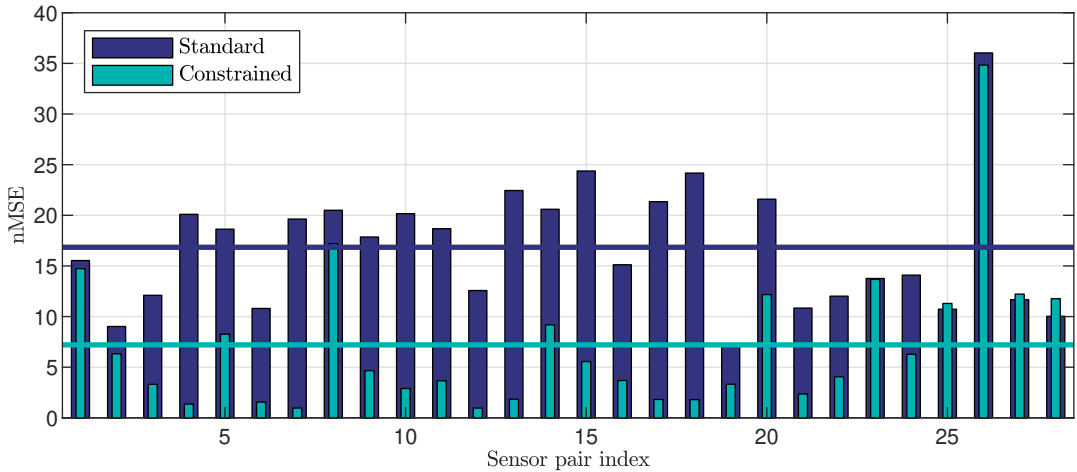}
    \caption{nMSE returned by each sensor pair model on a training set with limited coverage (case c). Horizontal lines correspond to the average of the standard and constrained nMSE across all sensor pairs.}
    \label{fig:4.2}
\end{figure}

To further investigate the performance of the constrained GP, we now consider just a single grid spacing of 30mm as an example, where no boundary measurements are available (case c in Figure \ref{fig:4.1d}). Case (c) was chosen as it presents the most challenging learning task. At a particular grid spacing, as there are a total of 28 sensor pair combinations, there are thus 28 total feature maps to learn. Figure \ref{fig:4.2} plots the nMSE obtained on the test set for each of the individual sensor pairing. For reference, \ref{appendix3} lists the sensor pair numbers with the corresponding index used here. 

\begin{figure}[h]
	\centering
	\includegraphics{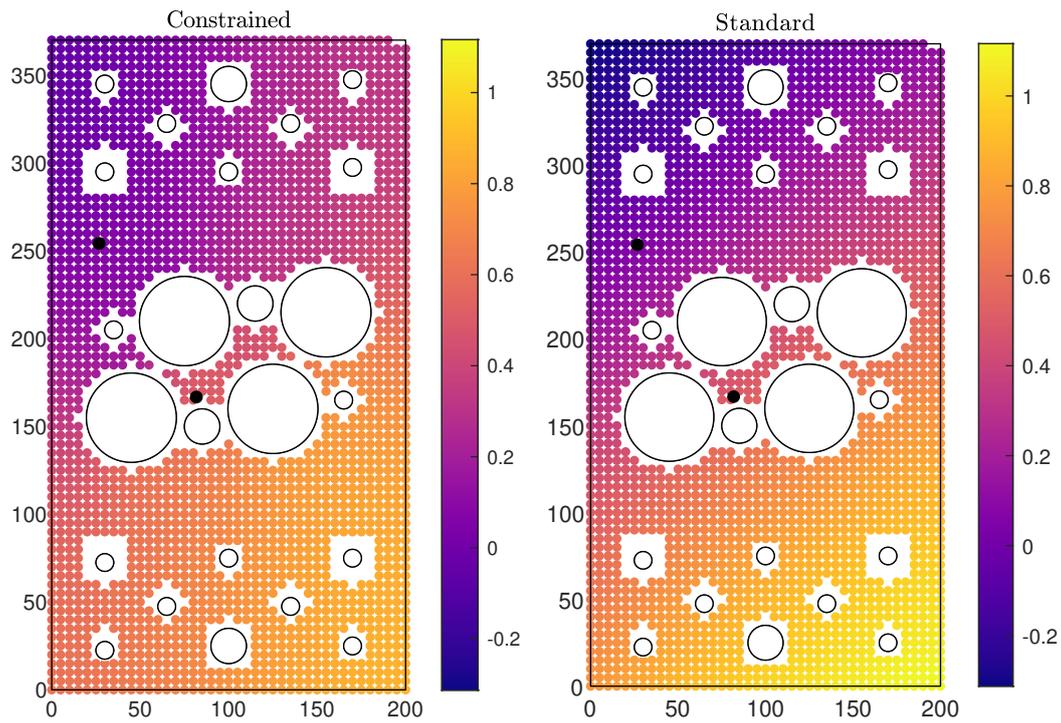}
	\caption{$\Delta$$T$ predictive mean of both GP models on the test set for sensor pair 3 \& 5. Sensor locations are highlighted by the black circles.}
    \label{fig:4.3}
  \end{figure}

As an initial observation, it can be seen that for the majority of the sensor pairs, the predictions returned by the constrained GP are vastly improved. This can be quantified formally by considering the averaged nMSE across all of the sensor pairs for both models, where the constrained GP yields an averaged error of 7.30 in comparison to 16.91 from the standard GP. Considering examples of the constrained GP performing significantly better, sensor pairing 15, which corresponds to sensors 3 \& 5, displays a large difference in error between the two models. Figure \ref{fig:4.3} maps the mean predictions on the test set for both of the GP models. Comparing both of the plots, it can be seen that most of the variation between the $\Delta T$ predictions exists in the upper and lower part of the plate, away from the location of the training data. Where predictions are made closer to the training points, such as in the centre of the plate, both models return similar predictions. 

\begin{figure}[h!]
	\centering
	\includegraphics{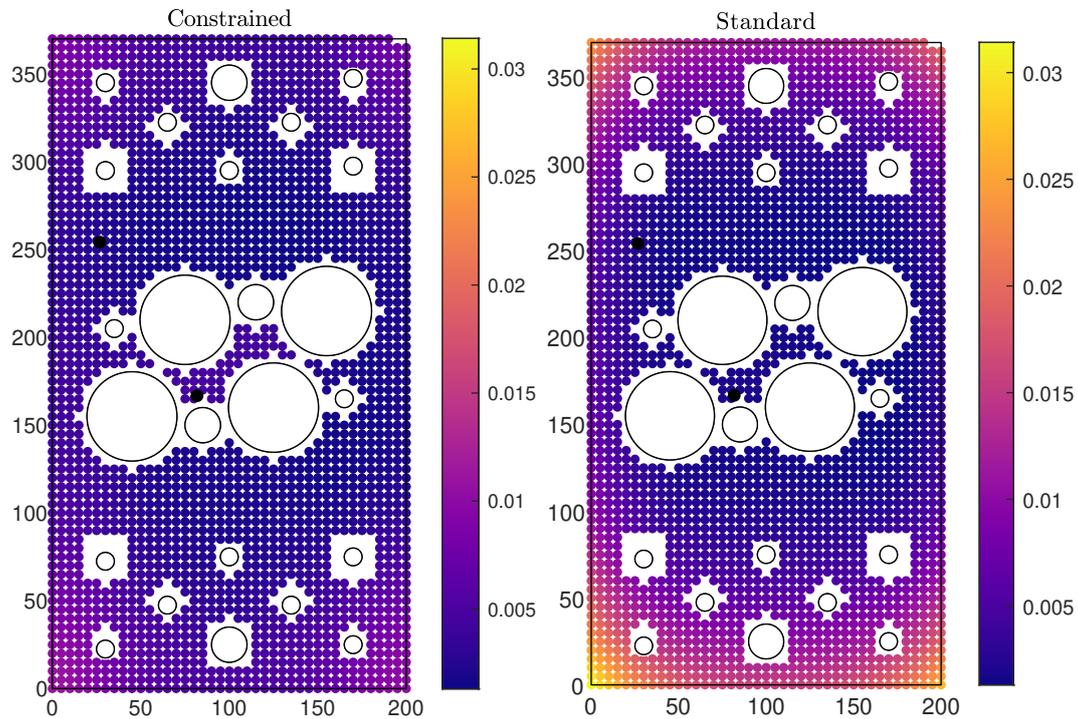}
	\caption{Predictive variance of both GP models for sensor pairing 3 \& 5. Sensor locations are highlighted by the black circles.}
    \label{fig:4.4}
\end{figure}

As the GP learns a distribution at each test location, as well as a predictive mean, a predictive variance is also computed. This allows uncertainty to be quantified across the prediction space, which is often very
desirable in SHM, offering a deeper level of insight to feed forward into assessments made regarding damage. Continuing with the analysis of sensor pair 3 \& 5, Figure \ref{fig:4.4} shows the predictive variance returned by both the standard and constrained GP on the test set. The results show that the constraints embedded into the GP generally reduce the uncertainty of the predictions made across the testing set compared to the standard GP, particularly in the upper and lower region of the plate. It is also possible for the predictive variance to be used to compute a log loss, providing a probabilistic error measure for the predictions made. However, as each of the training sets consist of uniformly spaced observations, differences in the predictive variance between the constrained and unconstrained model will generally only occur around the boundaries. When computing the log loss for each training scenario considered in this section so far, a similar trend will, therefore, be observed to that obtained for the nMSE. As such, log loss plots are not included in the main text, but for completeness, are provided in \ref{appendix4}.

% BELOW TO INCLUDE IN THESIS!!

%It is interesting to note that in the centre of the plate, the constraints do slightly increase the predictive variance. A possible explanation here is that the standard GP does not sufficiently represent the uncertainty in this region. Due to the close proximity of multiple boundaries in the centre of the plate, the onset time functions will vary more rapidly than in other areas of the plate; the degree of which becomes increasingly unknown as training points are reduced. Therefore, when training data are sparse, as considered in a 30mm spacing, the predictive variance should grow to reflect this additional uncertainty. For the standard GP, the scale at which the possible function draws vary is only controllable through the length scale hyperparameter, which is learnt to minimise the model evidence across the whole domain. There is, therefore, no mechanism for the standard GP to account for the uncertainty added by the increased sharpness of the $\Delta T$ functions in the middle of the plate. Where boundary constraints are included, the model has to satisfy the relevant conditions at the boundary locations. The possible onset function draws arising from the model are therefore aware of this sharper function behaviour, which is reflected by the increase in predictive variance. This functionality is important, and results in the constrained GP more reliably indicating regions of (un)certainty in the $\Delta T$ maps.

\begin{figure}[h]
    \centering
    \includegraphics{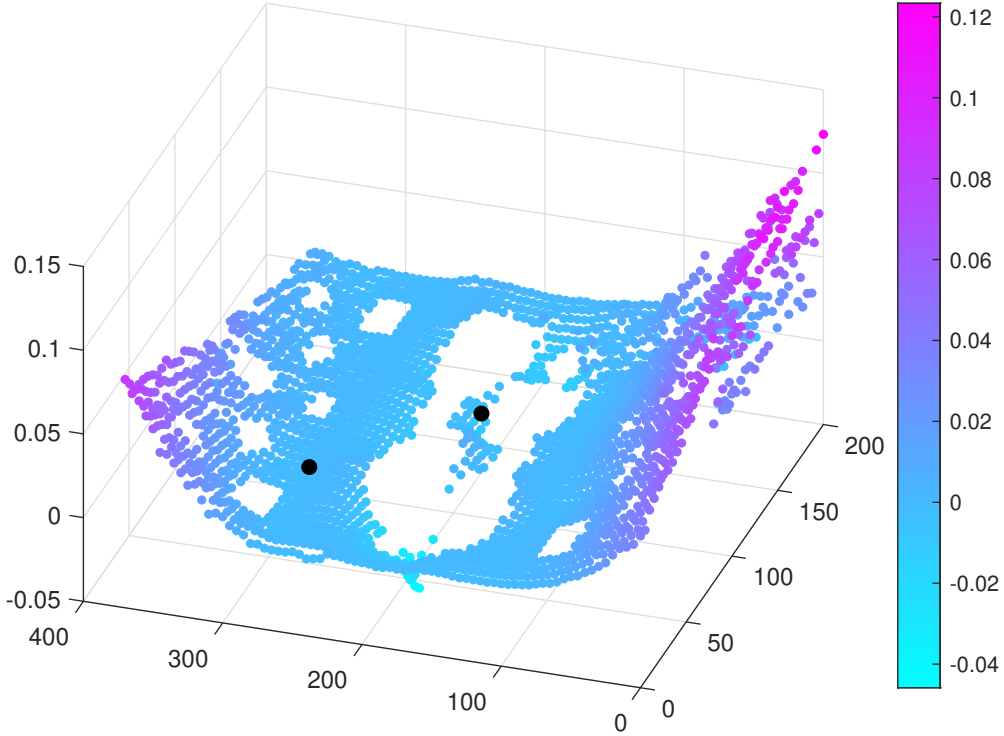}
    \caption{Squared error difference between standard and constrained GP for sensor pair 3 \& 5. Training observations
    taken from the middle section of the plate.} 
    \label{fig:4.5}
\end{figure}

Whilst these plots illustrate the learnt distribution over the AE features, they do not provide a direct measure of how well the predictions reflect the true target values. To analyse the discrepancy between the predicted and true targets for both model forms, a mapping of the squared error of the standard GP subtracted from the squared error of the constrained GP is considered. Under this metric, a positive value indicates that the error is larger in the standard GP, whilst negative values express a larger error in the constrained GP. Figure \ref{fig:4.5} maps this error metric across the test set for sensor pair 3 \& 5.

%Additionally, as the prediction points do not extend right up to the boundaries of the inner holes, did not extend right up to the boundaries of the inner holes,  in addition to where *no predictions were made   in the middle of the plate where 

The figure clearly demonstrates that in the upper and lower segments of the plate, which are the regions away from the training points, the accuracy of the $\Delta T$ predictions are improved by the physics-informed GP. As the prediction locations move further away from where the training data points are placed, the level of improvement offered by the constraints generally grows, particularly towards the boundaries at the upper and lower edges of the plate. For regions where training data coverage is good, then the use of boundary constraints will not significantly affect the mean predictions, explaining the similar error score obtained for both models in the centre of the plate. If the testing set contained data points on (or closer to) the inner hole boundaries, then it is likely that an error decrease would have been observed at these positions for the constrained model. However, testing measurements were only collected at a minimum distance of around 5mm from the inner holes.
A final observation that can be made is that the error reduction obtained in the upper left of the plate $(x=0,y=370)$ is greater than that returned in the upper right $(x=200,y=370)$, despite these two areas being geometrically symmetrical. To explain this behaviour, one must first recognise that the true $\Delta T$ values in a particular part of the plate will contain a level of variability that is dependent on the complexity of the propagation path between the locations of the sources and the receiving sensor. As the propagation path becomes more complex, whether that be due to the waves having to travel further to a receiving sensor, or because the propagation path is heavily obstructed (such as by multiple holes), then the variability of the onset times in a given region will increase, resulting in a more challenging feature map to learn in that part of the plate. For sensors 3 \& 5, the sensor pairing is positioned closer to the upper left of the plate than the right, requiring AE sources from the upper right region to propagate further to the receiving sensors, leading to a significant increase in the level of variability in the onset times in the upper right region than the upper left. As the constraints implemented here act only in relation to boundary conditions, it should not be expected that they provide a means of capturing this variability that arises from sensor positioning, and as such, both models perform similarly in this region. 

%One way to account for this behaviour is through the use of a heteroscedastic Gaussian process, which has been discussed in *. 

\begin{figure}[h]
    \centering
    \includegraphics{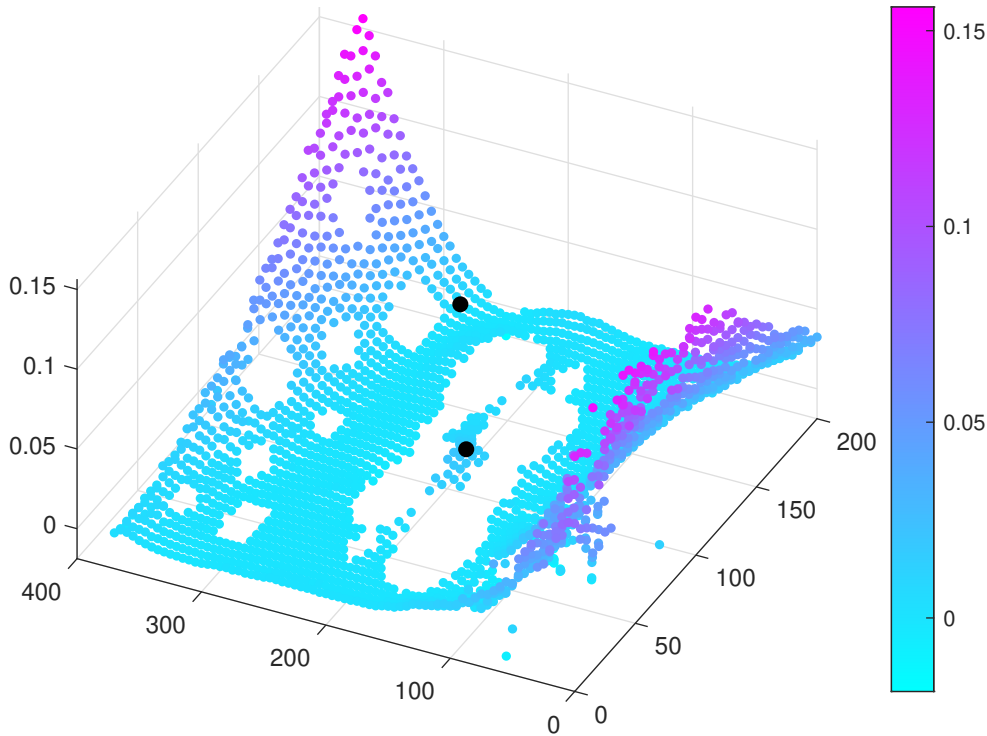}
    \caption{Squared error difference between standard and constrained GP for sensor pair 1 \& 5. Training observations taken from the middle section of the plate.} 
    \label{fig:4.6}
\end{figure}

Examining a second sensor pair mapping, Figure \ref{fig:4.6} plots the difference in square error for sensor pair 4 (sensors 1 \& 5). Again, it can be seen that away from the training data, the constraints significantly reduce the error of the $\Delta T$ predictions, with the maximum improvement occurring on the upper and lower boundaries of the plate. However, unlike Figure \ref{fig:4.5}, there is now a significant error reduction on the upper right of the plate, with the previously seen improvement on the upper left now absent. Again, the position of the sensor pair explains this behaviour, with sensor coverage improved in the top right of the plate, but reduced towards the upper left region. 

The dependance of the sensor coverage on the predictive performance also explains the reasoning for a number of sensor pairs returning a comparable error metric for the constrained GP with the standard GP. For example, sensors 1 \& 2 (index 1) and sensors 6 \& 7 (index 26). In these cases, the sensor coverage is poor, with both sensors generally lying adjacent to one another. This positioning results in large portions of the plate requiring waves to propagate further across the structure before being received, resulting in a more complex propagation path and, therefore, more variable $\Delta T$ features. If the interest is in improving predictions in locations with reduced sensor coverage, then implementing a constrained machine learner in isolation is not suitable. This is because constrained learners are still reliant on some baseline level of training data; for the constrained GP, the covariance structure of the features still needs to be learnt from input data, which are then used exclusively to make predictions. A potential approach to mitigate the effect of poor sensor coverage will be discussed in the following section, and forms a logical progression for future work.

\section{Conclusions}

This paper has demonstrated how known boundary conditions may be embedded into a Gaussian process regression model for learning acoustic emission onset time maps. For the time of arrival mapping problem, and also spatial modelling problems more generally, where there exists a lack of boundary measurements or restricted coverage of the total input space, it is shown that constraining the covariance function of the Gaussian process offers a significantly improved predictive performance. Due to the time and cost requirements of acquiring training data for real engineering structures, these are scenarios that consistently arise, illustrating the benefit of incorporating physical insight into the Gaussian process model through the approach presented in this paper.

Boundary conditions are just one example of insight obtained through an understanding of the underlying physics of the problem that may be harnessed when learning a Gaussian process model. In future work, additional engineering knowledge of the propagation behaviour of the acoustic emissions will be exploited to derive a physics-based mean function, with a view towards improved model performance in even more sparse training data regimes and where sensor coverage may be limited.

% good para:Given that the lengthier the data acquisition process is, the less likely asset mangers will be to seek the implementation of a health monitoring systems, these scenarios require

% In this work, embedding a Gaussian process with physical boundary conditions has
% been investigated from the perspective of learning dTOA maps. For spatial
% modelling problems where there exists a lack of boundary measurements or
% restricted coverage of the full input space, it is shown that constraining the
% covariance function of the GP prior offers an enchased predictive capability,
% both in mean predictions and uncertainty quantification. Due to the cost of
% acquiring training data for engineering structures, these are scenarios that
% consistently arise in SHM, and therefore illustrate the benefit of incorporating
% physical insight into the Gaussian process through boundary condition
% constraints. Further work will include other means for constraining Gaussian
% processes.

% Given that the constrained GP is aware of the process conditions that exist on the boundaries, this behaviour confirms that adding physical insight into the black-box learner improves predictive capability when less training information exists. 

% *** LIZZY COMMENT *** to include: In this case (i.e. bigger lack of data), one may incorporate a mean
%function derived from physical knowledge, and will be considered as part of future work

\section*{Acknowledgements}

The authors are grateful for the support of grant reference numbers EP/S001565/1 and EP/R004900/1. Thanks is offered to James Hensman, Mark Eaton, Robin Mills and Gareth Pierce for their work in the generating the data used throughout this paper. The support of Keith Worden during the completion of this work is also gratefully acknowledged. 

\setcounter{section}{0}
\setcounter{equation}{0}
\renewcommand\theequation{A.\arabic{equation}}
\renewcommand{\thesection}{Appendix \Alph{section}}

\section{Marginal likelihood}\label{app:appendix1}

To determine the hyperparameters, $\boldsymbol{\theta}$, of the covariance, a typical approach is to maximise the log marginal likelihood of the predictions with respect to $\boldsymbol{\theta}$. This is typically framed as a minimisation of the negative log marginal likelihood, which for the standard Gaussian process model, is expressed as,

\begin{equation}
    \boldsymbol{\theta} =  \argminA_{\boldsymbol{\theta}}\{\frac{1}{2}\mathbf{y}^T(K_{XX} +\sigma^2_n\mathbb{I})^{-1}\mathbf{y} + \frac{1}{2}\log |(K_{XX}+\sigma^2_n\mathbb{I})| + \frac{n}{2}\log(2\pi)\}.\label{eq:a.1}
    \end{equation}  

\noindent For the constrained GP, there exists a modified negative log marginal likelihood, 

\begin{equation}\label{eq:a.2}
    \begin{aligned}
        \boldsymbol{\theta} =  \argminA_{\boldsymbol{\theta}}\{\frac{1}{2}(N-m)\log\sigma_n^2&+\frac{1}{2}\sum_{j=1}^m\boldsymbol{\Lambda}_{j,j} + \frac{1}{2}\log|\sigma_n^2\boldsymbol{\Lambda}^{-1} + \boldsymbol{\Phi}^T\boldsymbol{\Phi}|+\frac{n}{2}\log(2\pi) \\ &+
         \frac{1}{2\boldsymbol{\sigma}_n^2}(\mathbf{y}^T\mathbf{y}-\mathbf{y}^T\boldsymbol{\Phi}(\sigma_n^2\boldsymbol{\Lambda}^{-1}+\boldsymbol{\Phi}^T\boldsymbol{\Phi})^{-1}\boldsymbol{\Phi}\mathbf{y})\}.
         \end{aligned}
 \end{equation}

\noindent For optimising both the standard and constrained covariance hyperparameters, a Quantum-behaved Particle Swarm Optimisation (QPSO) \cite{sun2004global,worden2018evolutionary} is employed, a gradient-free heuristic optimiser.

\setcounter{equation}{0}
\renewcommand\theequation{B.\arabic{equation}}
\setcounter{figure}{0}
\renewcommand\thefigure{B.\arabic{figure}}

\section{Laplace approximation with derivative boundary conditions}\label{appendix2}

\begin{figure}[h]\label{fig:3.4}
    \centering
    \subfloat[Numerical grid]{\includegraphics[scale=0.225]{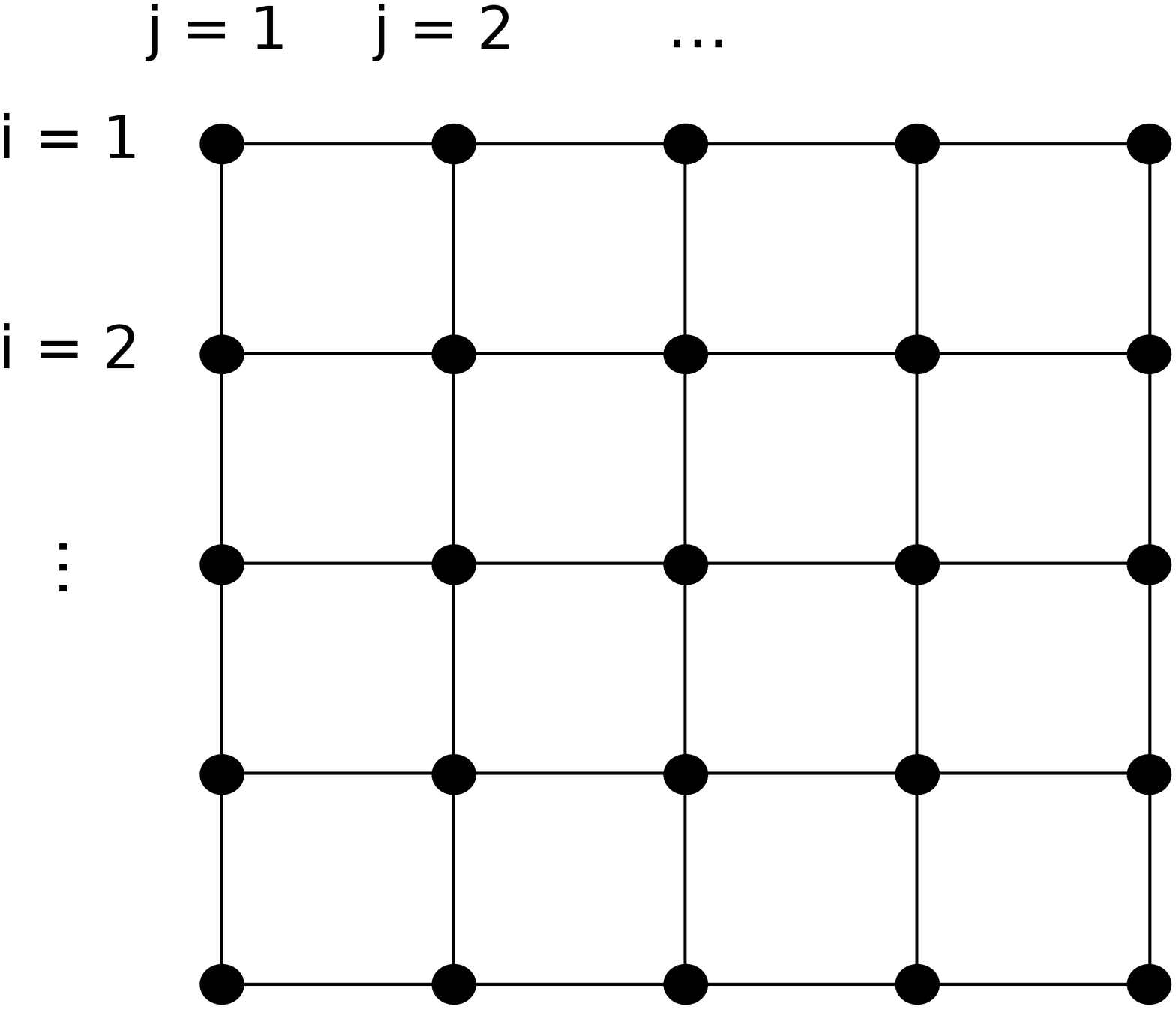}\label{fig:3.4a}}
    \hspace{5mm}
    \subfloat[Numerical grid with ghost points]{\includegraphics[scale=0.225]{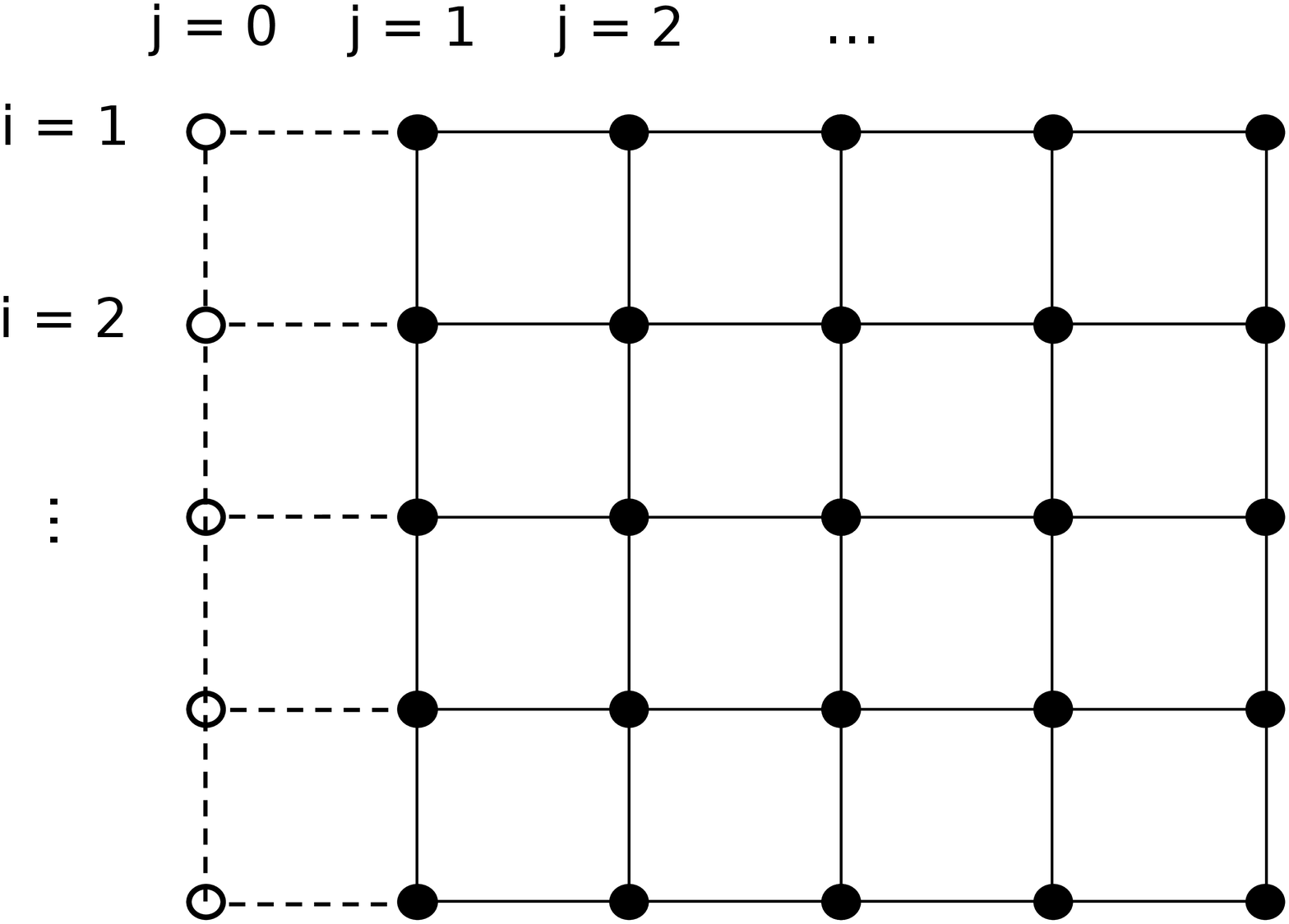}\label{fig:3.4b}}
    \caption{Numerical grid for arbitrary domain to depict how ghost points are introduced to include derivative boundary conditions.}
  \end{figure}

Solving for first order derivative boundary conditions is slightly more involved than Dirichlet conditions and requires the use of a first order approximation at the boundary location. However, before incorporating this boundary type, a series of ``ghost points" must be added to every location directly adjacent to a boundary. These ghosts points are purely artificial, and exist to help implement the finite difference approximation at endpoints of the domain. As an example, take the arbitrary 2D domain shown in Figure \ref{fig:3.4a}, where each circle represents a node in the grid mask. 
Consideration of the boundary on the left hand side of the domain along the $j$ direction requires the addition of ghost points at $j=0$, as shown in Figure \ref{fig:3.4b}. 

Incorporation of a first order derivative at $j=1$ can then be achieved by applying a backward difference approximation 

\begin{equation} \label{eq:3:3}
    u_j' = \frac{u_{i,j} -u_{i,j-1}}{h},
\end{equation}

%\begin{equation} \label{eq:3.3}
    %u_i' = \frac{u_{i+1,j} - u_{i,j}}{h}, \hspace{5mm} u_j' = \frac{u_{i,j+1} -u_{i,j}}{h},
%\end{equation}

\noindent which for zero derivative boundary conditions, can be simplified to,

\begin{equation}\label{eq:3.4}
    u_{i,j} = u_{i,j-1}.
\end{equation}

\noindent At this point, it is worth noting that although a higher order approximation would be obtained through the use of a central difference method, the use of a backward (and forward, as discussed shortly) difference scheme ensures that the resultant stencil matrix is symmetric \cite{leveque2007finite}, and so the corresponding eigenvectors are real. Returning to equation (\ref{eq:3.1}), at $j=1$, the ghost points that appear at $u_{i,0}$ can be removed by substituting in the above expression, yielding,

\begin{equation}\label{eq:3:5}
    -\nabla^2u(i,1) \approx \frac{1}{h^2}(-3u_{i,1} + u_{i-1,1} + u_{i,2}),
\end{equation}

\noindent and incorporating the boundaries into the stencil matrix. Where a ghost point lies at an index of $+1$ to the boundary, such as the right hand side of the domain in Figure \ref{fig:3.4a}, then it is necessary for a forward difference approximation to be applied

\begin{equation}\label{eq:3:6}
    u_j' = \frac{u_{i,j+1} -u_{i,j}}{h},
\end{equation}

\noindent replacing equation (\ref{eq:3.4}) with, 

\begin{equation} \label{eq:3.7}
    u_{i,j+1} = u_{i,j}
\end{equation}

\noindent The process detailed above is then identical for any given direction, providing the finite difference equation applied is consistent with the target direction. For example, in the $i$ direction, equations (\ref{eq:3.4}) and (\ref{eq:3.7}) become

\begin{align} \label{eq:3.8}
    u_{i,j} &= u_{i-1,j} \\
    u_{i,j} &= u_{i+1,j} \label{eq:3.9}.
\end{align}

\noindent Iterating through each element of the grid mask, a stencil matrix can be computed that corresponds to the negative Laplacian of $\Omega$.
The leading $m$ eigenvalues and eigenfunctions can then be calculated through a chosen numerical solver. 

\setcounter{equation}{0}
\renewcommand\theequation{C.\arabic{equation}}
\setcounter{table}{0}
\renewcommand\thetable{C.\arabic{table}}

\newpage

\section{Sensor pair index}\label{appendix3}

\begin{table}[h!]
    \centering
        \begin{tabular}{cccc}
        \hline
        Sensor pair index & Sensors & Sensor pair index & Sensors \\ \hline
        1                 & 1-2     & 15                & 3-5     \\
        2                 & 1-3     & 16                & 3-6     \\
        3                 & 1-4     & 17                & 3-7     \\
        4                 & 1-5     & 18                & 3-8     \\
        5                 & 1-6     & 19                & 4-5     \\
        6                 & 1-7     & 20                & 4-6     \\
        7                 & 1-8     & 21                & 4-7     \\
        8                 & 2-3     & 22                & 4-8     \\
        9                 & 2-4     & 23                & 5-6     \\
        10                & 2-5     & 24                & 5-7     \\
        11                & 2-6     & 25                & 5-8     \\
        12                & 2-7     & 26                & 6-7     \\
        13                & 2-8     & 27                & 6-8     \\
        14                & 3-4     & 28                & 7-8    
        \end{tabular}
    \caption{Sensor pair index and corresponding individual sensors.}
    \end{table}

\setcounter{equation}{0}
\renewcommand\theequation{C.\arabic{equation}}
\setcounter{table}{0}
\renewcommand\thetable{D.\arabic{table}}

\section{Log loss results}\label{appendix4}

The mean standardised log loss (MSLL) is expressed as,  

\begin{equation}
    MSLL = \frac{1}{N}\sum_k\{-\log p(\mathbf{y}_{\star,k}| X,\mathbf{y},\mathbf{x}_{\star,k}) + \log p(\mathbf{y}_{\star,k};\mathbb{E}(\mathbf{y}_k), \mathbb{V}(\mathbf{y}_k))\},
\end{equation}

\noindent where $k$ indexes a particular test point. The log loss can be interpreted as the negative likelihood of the predictions relative to those made under the trivial model, i.e. the mean and variance of the training observations. As such, a larger negative MSLL reflects more favourable predictive distributions.

For each of the training scenarios considered in Section \ref{sec:4} (e.g. Figure \ref{fig:4.1a}-\ref{fig:4.1d}), equivalent MSLL plots are provided below. 

\begin{figure}[h!]
    \includegraphics{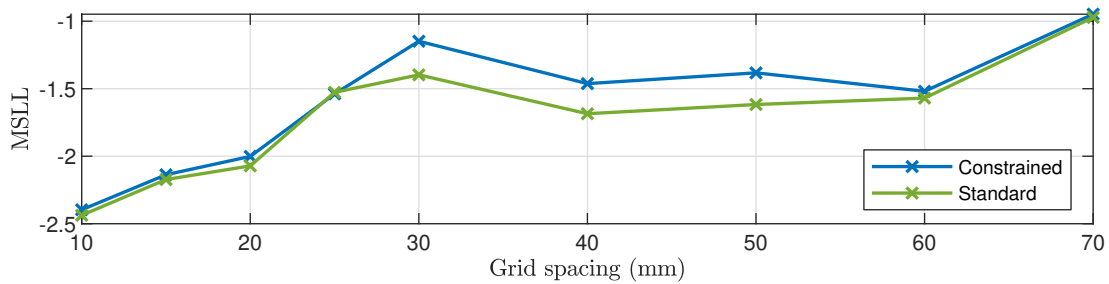}
    \caption{MSLL of the predictions made by the constrained and standard Gaussian process as the number of training points are reduced. Here the training points fully cover the plate, but at different grid densities.}
    \label{fig:D1}
\end{figure}

\begin{figure}[h!]
    \includegraphics{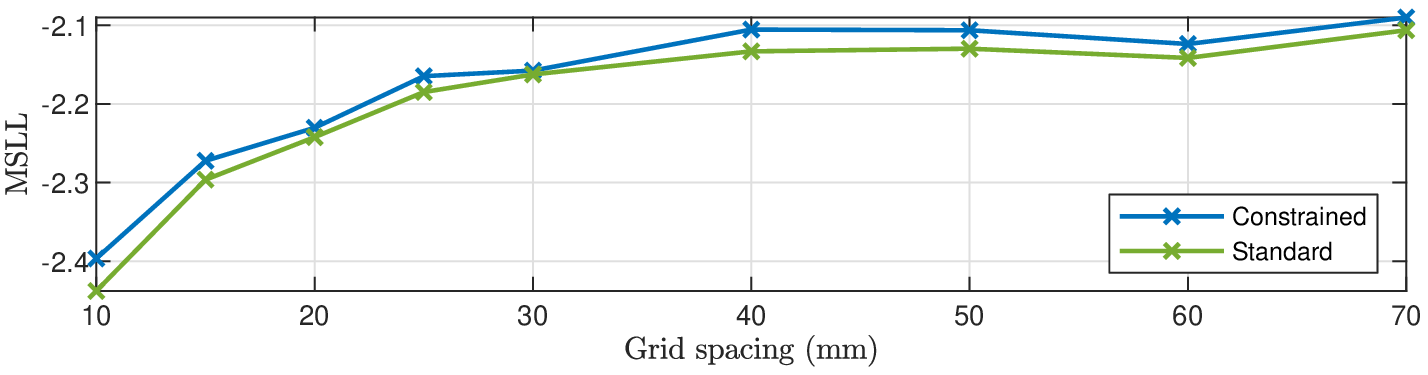}
    \caption{MSLL of the predictions made by the constrained and standard Gaussian process as training points are reduced whilst retaining full boundary measurements.}
    \label{fig:D2}
\end{figure}

\begin{figure}[h!]
    \includegraphics{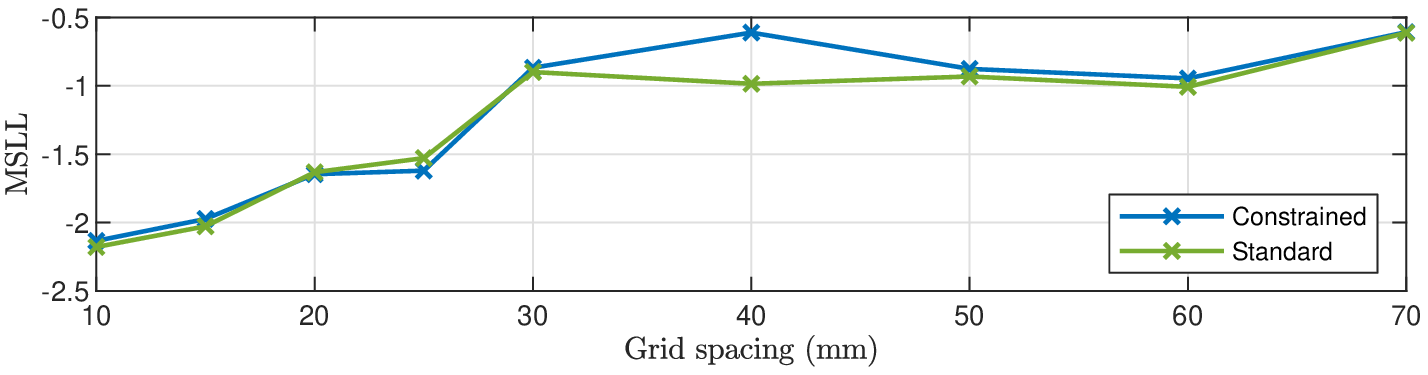}
    \caption{MSLL of the predictions made by the constrained and standard Gaussian process as training points are reduced with no boundary measurements.}
    \label{fig:D3}
\end{figure}

\begin{figure}[h!]
    \includegraphics{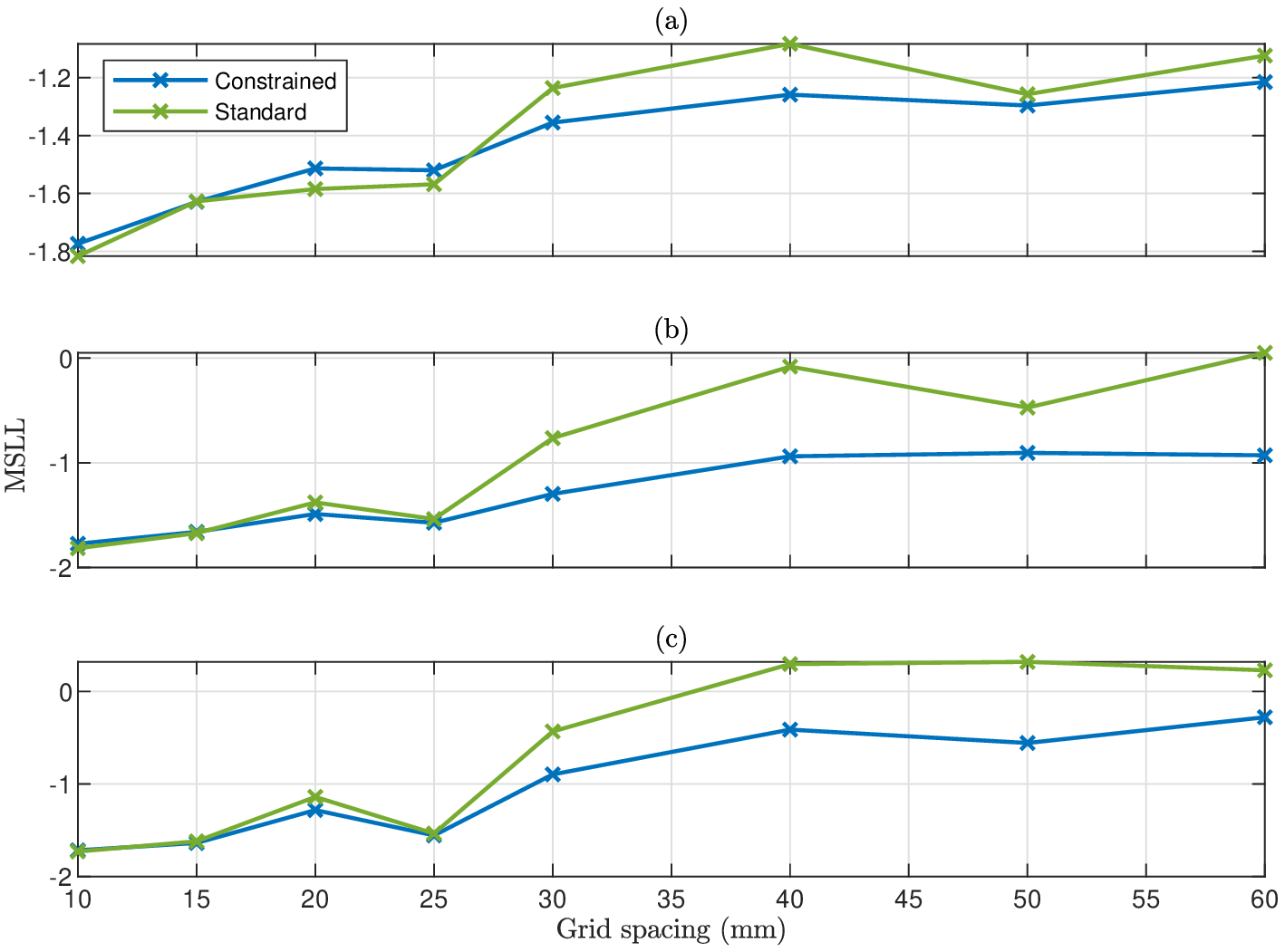}%
    \caption{MSLL with restricted training coverage for a) full boundary location measurements b) partial boundary measurements c) no boundary measurements.}
    \label{fig:D4}
\end{figure}

\newpage

\bibliographystyle{IEEEtran.bst}
\bibliography{references.bib}

\begin{thebibliography}{10}
\providecommand{\url}[1]{#1}
\csname url@samestyle\endcsname
\providecommand{\newblock}{\relax}
\providecommand{\bibinfo}[2]{#2}
\providecommand{\BIBentrySTDinterwordspacing}{\spaceskip=0pt\relax}
\providecommand{\BIBentryALTinterwordstretchfactor}{4}
\providecommand{\BIBentryALTinterwordspacing}{\spaceskip=\fontdimen2\font plus
\BIBentryALTinterwordstretchfactor\fontdimen3\font minus
  \fontdimen4\font\relax}
\providecommand{\BIBforeignlanguage}[2]{{%
\expandafter\ifx\csname l@#1\endcsname\relax
\typeout{** WARNING: IEEEtran.bst: No hyphenation pattern has been}%
\typeout{** loaded for the language `#1'. Using the pattern for}%
\typeout{** the default language instead.}%
\else
\language=\csname l@#1\endcsname
\fi
#2}}
\providecommand{\BIBdecl}{\relax}
\BIBdecl

\bibitem{farrar2012structural}
C.~R. Farrar and K.~Worden, \emph{Structural health monitoring: a machine
  learning perspective}.\hskip 1em plus 0.5em minus 0.4em\relax John Wiley \&
  Sons, 2012.

\bibitem{rogers2019bayesian}
T.~Rogers, K.~Worden, R.~Fuentes, N.~Dervilis, U.~Tygesen, and E.~Cross, ``A
  {B}yesian non-parametric clustering approach for semi-supervised structural
  health monitoring,'' \emph{Mechanical Systems and Signal Processing}, vol.
  119, pp. 100--119, 2019.

\bibitem{bull2018active}
L.~Bull, K.~Worden, G.~Manson, and N.~Dervilis, ``Active learning for
  semi-supervised structural health monitoring,'' \emph{Journal of Sound and
  Vibration}, vol. 437, pp. 373--388, 2018.

\bibitem{avendano2017gaussian}
L.~D. Avenda{\~n}o-Valencia, E.~N. Chatzi, K.~Y. Koo, and J.~M. Brownjohn,
  ``Gaussian process time-series models for structures under operational
  variability,'' \emph{Frontiers in Built Environment}, vol.~3, p.~69, 2017.

\bibitem{cross2012}
E.~J. Cross, ``On structural health monitoring in changing environmental and
  operational conditions,'' Ph.D. dissertation, University of Sheffield, 2012.

\bibitem{tobias1976acoustic}
A.~Tobias, ``Acoustic-emission source location in two dimensions by an array of
  three sensors,'' \emph{Non-destructive testing}, vol.~9, no.~1, pp. 9--12,
  1976.

\bibitem{hensman2010locating}
J.~Hensman, R.~Mills, S.~Pierce, K.~Worden, and M.~Eaton, ``Locating acoustic
  emission sources in complex structures using {G}aussian processes,''
  \emph{Mechanical Systems and Signal Processing}, vol.~24, no.~1, pp.
  211--223, 2010.

\bibitem{ciampa2012impact}
F.~Ciampa and M.~Meo, ``Impact detection in anisotropic materials using a time
  reversal approach,'' \emph{Structural Health Monitoring}, vol.~11, no.~1, pp.
  43--49, 2012.

\bibitem{niri2014nonlinear}
E.~D. Niri, A.~Farhidzadeh, and S.~Salamone, ``Nonlinear kalman filtering for
  acoustic emission source localization in anisotropic panels,''
  \emph{Ultrasonics}, vol.~54, no.~2, pp. 486--501, 2014.

\bibitem{kundu2014acoustic}
T.~Kundu, ``Acoustic source localization,'' \emph{Ultrasonics}, vol.~54, no.~1,
  pp. 25--38, 2014.

\bibitem{jones2020bayesian}
M.~R. Jones, T.~J. Rogers, K.~Worden, and E.~J. Cross, ``A bayesian methodology
  for localising acoustic emission sources in complex structures,''
  \emph{Mechanical Systems and Signal Processing}, vol. 163, p. 108143, 2022.

\bibitem{baxter2007delta}
M.~G. Baxter, R.~Pullin, K.~M. Holford, and S.~L. Evans, ``Delta {T} source
  location for acoustic emission,'' \emph{Mechanical systems and signal
  processing}, vol.~21, no.~3, pp. 1512--1520, 2007.

\bibitem{pearson2017improved}
M.~R. Pearson, M.~Eaton, C.~Featherston, R.~Pullin, and K.~Holford, ``Improved
  acoustic emission source location during fatigue and impact events in
  metallic and composite structures,'' \emph{Structural Health Monitoring},
  vol.~16, no.~4, pp. 382--399, 2017.

\bibitem{ebrahimkhanlou2018single}
A.~Ebrahimkhanlou and S.~Salamone, ``Single-sensor acoustic emission source
  localization in plate-like structures using deep learning,''
  \emph{Aerospace}, vol.~5, no.~2, p.~50, 2018.

\bibitem{willard2020integrating}
J.~Willard, X.~Jia, S.~Xu, M.~Steinbach, and V.~Kumar, ``Integrating
  physics-based modeling with machine learning: {A} survey,'' \emph{arXiv
  preprint arXiv:2003.04919}, 2020.

\bibitem{cross2022physics}
E.~J. Cross, S.~J. Gibson, M.~R. Jones, D.~J. Pitchforth, S.~Zhang, and T.~J.
  Rogers, \emph{Structural Health Monitoring based on Data Science
  Techniques}.\hskip 1em plus 0.5em minus 0.4em\relax Springer - IN PRESS,
  2022, ch. Physics-informed machine learning for Structural Health Monitoring.

\bibitem{williams2006gaussian}
C.~K. Williams and C.~E. Rasmussen, \emph{Gaussian {P}rocesses for {M}achine
  {L}earning}.\hskip 1em plus 0.5em minus 0.4em\relax MIT press Cambridge, MA,
  2006, no.~3.

\bibitem{swiler2020survey}
L.~P. Swiler, M.~Gulian, A.~L. Frankel, C.~Safta, and J.~D. Jakeman, ``A survey
  of constrained {G}aussian process regression: {A}pproaches and implementation
  challenges,'' \emph{Journal of Machine Learning for Modeling and Computing},
  vol.~1, no.~2, 2020.

\bibitem{agrell2019gaussian}
C.~Agrell, ``Gaussian processes with linear operator inequality constraints,''
  \emph{arXiv preprint arXiv:1901.03134}, 2019.

\bibitem{riihimaki2010gaussian}
J.~Riihim{\"a}ki and A.~Vehtari, ``Gaussian processes with monotonicity
  information,'' in \emph{Proceedings of the thirteenth international
  conference on artificial intelligence and statistics}.\hskip 1em plus 0.5em
  minus 0.4em\relax JMLR Workshop and Conference Proceedings, 2010, pp.
  645--652.

\bibitem{da2012gaussian}
S.~Da~Veiga and A.~Marrel, ``Gaussian process modeling with inequality
  constraints,'' in \emph{Annales de la Facult{\'e} des sciences de Toulouse:
  Math{\'e}matiques}, vol.~21, no.~3, 2012, pp. 529--555.

\bibitem{maatouk2017finite}
H.~Maatouk, ``Finite-dimensional approximation of {G}aussian processes with
  inequality constraints,'' \emph{arXiv preprint arXiv:1706.02178}, 2017.

\bibitem{lopez2018finite}
A.~F. L{\'o}pez-Lopera, F.~Bachoc, N.~Durrande, and O.~Roustant,
  ``Finite-dimensional {G}aussian approximation with linear inequality
  constraints,'' \emph{SIAM/ASA Journal on Uncertainty Quantification}, vol.~6,
  no.~3, pp. 1224--1255, 2018.

\bibitem{pensoneault2020nonnegativity}
A.~Pensoneault, X.~Yang, and X.~Zhu, ``Nonnegativity-enforced {G}aussian
  process regression,'' \emph{Theoretical and Applied Mechanics Letters},
  vol.~10, no.~3, pp. 182--187, 2020.

\bibitem{jidling2017linearly}
C.~Jidling, N.~Wahlstr{\"o}m, A.~Wills, and T.~B. Sch{\"o}n, ``Linearly
  constrained {G}aussian processes,'' \emph{arXiv preprint arXiv:1703.00787},
  2017.

\bibitem{cross2021physics}
E.~J. Cross and T.~J. Rogers, ``Physics-derived covariance functions for
  machine learning in structural dynamics,'' \emph{IFAC-PapersOnLine}, vol.~54,
  no.~7, pp. 168--173, 2021.

\bibitem{alvarez2009latent}
M.~Alvarez, D.~Luengo, and N.~D. Lawrence, ``Latent force models,'' in
  \emph{Artificial Intelligence and Statistics}.\hskip 1em plus 0.5em minus
  0.4em\relax PMLR, 2009, pp. 9--16.

\bibitem{raissi2018hidden}
M.~Raissi and G.~E. Karniadakis, ``Hidden physics models: Machine learning of
  nonlinear partial differential equations,'' \emph{Journal of Computational
  Physics}, vol. 357, pp. 125--141, 2018.

\bibitem{cross2019grey}
E.~J. Cross, T.~J. Rogers, and T.~J. Gibbons, ``Grey-box modelling for
  structural health monitoring: physical constraints on machine learning
  algorithms,'' \emph{Structural Health Monitoring 2019}, 2019.

\bibitem{solin2020hilbert}
A.~Solin and S.~S{\"a}rkk{\"a}, ``Hilbert space methods for reduced-rank
  {G}aussian process regression,'' \emph{Statistics and Computing}, vol.~30,
  no.~2, pp. 419--446, 2020.

\bibitem{stein2012interpolation}
M.~L. Stein, \emph{Interpolation of Spatial Data: Some Theory for
  Kriging}.\hskip 1em plus 0.5em minus 0.4em\relax Springer Science \& Business
  Media, 2012.

\bibitem{chatfield2019analysis}
C.~Chatfield and H.~Xing, \emph{The Analysis of Time Series: An Introduction
  with R}.\hskip 1em plus 0.5em minus 0.4em\relax CRC press, 2019.

\bibitem{pitchforth2021grey}
D.~Pitchforth, T.~Rogers, U.~Tygesen, and E.~Cross, ``Grey-box models for wave
  loading prediction,'' \emph{Mechanical Systems and Signal Processing}, vol.
  159, p. 107741, 2021.

\bibitem{zhang2020gaussian}
S.~Zhang, T.~J. Rogers, and E.~J. Cross, ``Gaussian process based grey-box
  modelling for shm of structures under fluctuating environmental conditions,''
  in \emph{European Workshop on Structural Health Monitoring}.\hskip 1em plus
  0.5em minus 0.4em\relax Springer, 2020, pp. 55--66.

\bibitem{al2016acoustic}
S.~K. Al-Jumaili, M.~R. Pearson, K.~M. Holford, M.~J. Eaton, and R.~Pullin,
  ``Acoustic emission source location in complex structures using full
  automatic {D}elta {T} mapping technique,'' \emph{Mechanical Systems and
  Signal Processing}, vol.~72, pp. 513--524, 2016.

\bibitem{jones2021bayesian}
M.~R. Jones, T.~J. Rogers, I.~E. Martinez, and E.~J. Cross, ``Bayesian
  localisation of acoustic emission sources for wind turbine bearings,'' in
  \emph{Health Monitoring of Structural and Biological Systems XV}, vol.
  11593.\hskip 1em plus 0.5em minus 0.4em\relax International Society for
  Optics and Photonics, 2021, p. 115932D.

\bibitem{solin2019know}
A.~Solin and M.~Kok, ``Know your boundaries: Constraining gaussian processes by
  variational harmonic features,'' in \emph{The 22nd International Conference
  on Artificial Intelligence and Statistics}.\hskip 1em plus 0.5em minus
  0.4em\relax PMLR, 2019, pp. 2193--2202.

\bibitem{jones2020isma}
M.~R. Jones, T.~J. Rogers, P.~Gardner, and E.~J. Cross, ``Constraining
  {G}aussian processes for grey-box acoustic emission source localisation,'' in
  \emph{ISMA 2020 - International Conference on Noise and Vibration Engineering
  and USD 2020 - International Conference on Uncertainty in Structural
  Dynamics}, 2020.

\bibitem{sun2004global}
J.~Sun, W.~Xu, and B.~Feng, ``A global search strategy of quantum-behaved
  particle swarm optimization,'' in \emph{IEEE Conference on Cybernetics and
  Intelligent Systems, 2004.}, vol.~1.\hskip 1em plus 0.5em minus 0.4em\relax
  IEEE, 2004, pp. 111--116.

\bibitem{worden2018evolutionary}
K.~Worden, R.~Barthorpe, E.~Cross, N.~Dervilis, G.~Holmes, G.~Manson, and
  T.~Rogers, ``On evolutionary system identification with applications to
  nonlinear benchmarks,'' \emph{Mechanical Systems and Signal Processing}, vol.
  112, pp. 194--232, 2018.

\bibitem{leveque2007finite}
R.~J. LeVeque, \emph{Finite difference methods for ordinary and partial
  differential equations: steady-state and time-dependent problems}.\hskip 1em
  plus 0.5em minus 0.4em\relax SIAM, 2007.

\end{thebibliography}
\end{document}